\documentclass[letterpaper, 10 pt, conference]{ieeeconf}  %

\IEEEoverridecommandlockouts                              %

\usepackage{amsfonts}       %
\usepackage{amsmath,amssymb}
\usepackage{graphicx}
\makeatletter
\let\NAT@parse\undefined
\makeatother
\usepackage[colorlinks]{hyperref}
\usepackage[capitalise]{cleveref}
\usepackage{multirow}
\usepackage[export]{adjustbox}

\usepackage{graphbox}
\usepackage{booktabs}
\usepackage{url}

\newcommand{\algname}{\mbox{HAICU}}
\newcommand{\emphalgname}{\emph{\algname}}
\newcommand{\dataname}{\mbox{PUP}}
\newcommand{\emphdataname}{\emph{\dataname}}

\newcommand\Tstrut{\rule{0pt}{2.6ex}}       %
\newcommand\Bstrut{\rule[-0.9ex]{0pt}{0pt}} %
\newcommand{\TBstrut}{\Tstrut\Bstrut} %

\usepackage{url}

\renewcommand{\baselinestretch}{0.95}

\title{\LARGE \bf
Heterogeneous-Agent Trajectory Forecasting\\Incorporating Class Uncertainty
}

\author{Boris Ivanovic$^{1\dagger}$\hspace{1cm} Kuan-Hui Lee$^{2}$\hspace{1cm} Pavel Tokmakov$^{2}$\hspace{1cm} Blake Wulfe$^{2}$\\Rowan McAllister$^{2}$\hspace{1cm} Adrien Gaidon$^{2}$\hspace{1cm} Marco Pavone$^{1,3}$%
\thanks{*We thank Jie Li for her input throughout the project. Toyota Research Institute (``TRI'') provided funds to assist the authors with their research but
this article solely reflects the opinions and conclusions of
its authors and not TRI or any other Toyota entity. We also acknowledge the support of the Natural Sciences and Engineering Research Council of Canada (NSERC), funding reference number 545934-2020.}%
\thanks{$^\dagger$This work was completed while the author was at Stanford University.}%
\thanks{$^1$Boris Ivanovic is with NVIDIA Research {\tt\small \{bivanovic@nvidia.com\}}}%
\thanks{$^2$Kuan-Hui Lee, Pavel Tokmakov, Blake Wulfe, Rowan McAllister, and Adrien Gaidon are with the Toyota Research Institute {\tt\small \{first.last\}@tri.global}}%
\thanks{$^3$Marco Pavone is with the Department of Aeronautics and Astronautics, Stanford University, and with NVIDIA Research {\tt\small \{pavone@stanford.edu, mpavone@nvidia.com\}}}%
}

\begin{document}

\maketitle
\thispagestyle{empty}
\pagestyle{empty}

\begin{abstract}
Reasoning about the future behavior of other agents is critical to safe robot navigation. The multiplicity of plausible futures is further amplified by the uncertainty inherent to agent state estimation from data, including positions, velocities, and semantic class. Forecasting methods, however, typically neglect class uncertainty, conditioning instead only on the agent's most likely class, even though perception models often return full class distributions. To exploit this information, we present \emphalgname{}, a method for heterogeneous-agent trajectory forecasting that explicitly incorporates agents' class probabilities. We additionally present \emphdataname{}, a new challenging real-world autonomous driving dataset, to investigate the impact of Perceptual Uncertainty in Prediction. It contains challenging crowded scenes with unfiltered agent class probabilities that reflect the long-tail of current state-of-the-art perception systems. We demonstrate that incorporating class probabilities in trajectory forecasting significantly improves performance in the face of uncertainty, and enables new forecasting capabilities such as counterfactual predictions.
\end{abstract}

\section{INTRODUCTION}

\begin{figure}[t]
    \centering
    \includegraphics[trim={0 3em 4em 3em},clip,width=0.5\linewidth,frame]{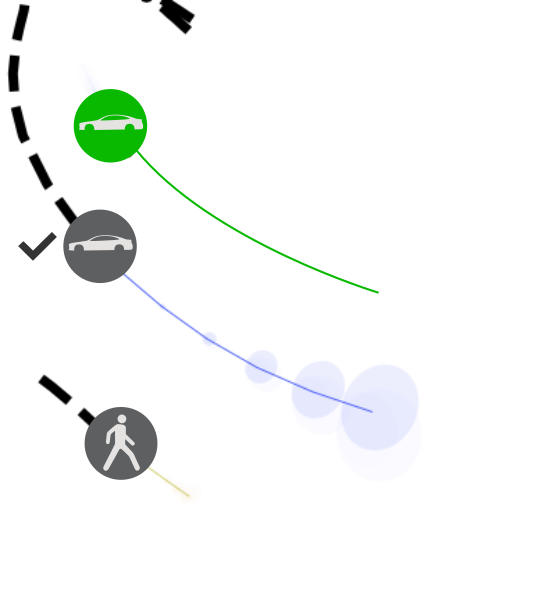}
    \includegraphics[trim={0 3.15em 4em 3.15em},clip,width=0.47\linewidth,frame]{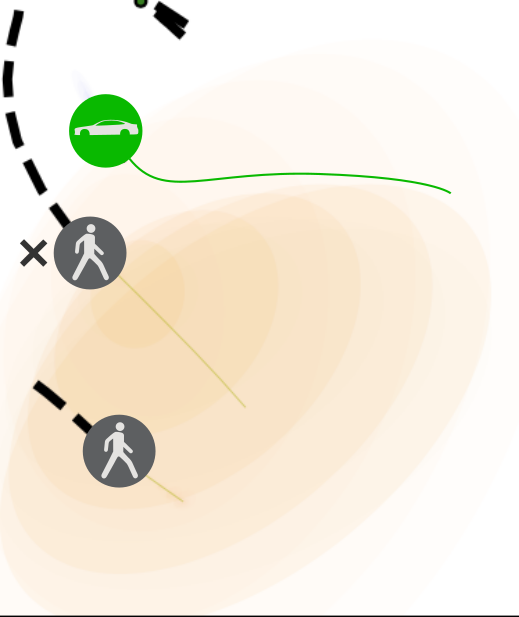}
    \caption{
    Classifying agents with their most-likely class can be disastrous in the presence of uncertainty. A state-of-the-art trajectory forecasting model, Trajectron++ \cite{SalzmannIvanovicEtAl2020}, produces well-behaved predictions when agents are correctly classified (\textbf{left}; as a car). However, its uncertainty grows sharply when the agent is misclassified (\textbf{right}; as a pedestrian). This can cause a sudden and unexpected change in the ego-vehicle's resulting motion plan, in green.
    }
    \label{fig:problem}
\vspace*{-0.5cm}
\end{figure}

Incorporating perceptual uncertainty into downstream components, such as forecasting and planning, is critical for the safe operation of autonomous vehicles.
However, most trajectory forecasting methods do not explicitly incorporate or propagate perceptual uncertainties from their inputs to their outputs~\cite{RudenkoPalmieriEtAl2019}. Instead, they classify an agent with its highest probability class from the upstream perception system. Although, blindly trusting the most-likely class can have disastrous consequences, especially in the presence of class uncertainty, as shown in \cref{fig:problem}. 
Misclassifications like these could cause an autonomous vehicle to make unnecessary evasive maneuvers and may occur frequently in challenging real-world conditions (\cref{sec:datasets}).
A safer way to deal with perceptual uncertainties is to propagate them through forecasting systems, so that planning components can make uncertainty-aware decisions~\cite{mcallister2017concrete,bhatt2020probabilistic}.

{\bf Contributions.} Our key contributions are threefold. 
First, we present \textbf{\algname{}}, a method for Heterogeneous-Agent trajectory forecasting Incorporating Class Uncertainty (\cref{sec:method}). We show that directly incorporating class probabilities from upstream perception systems into a state-of-the-art trajectory forecasting method effectively improves performance in the presence of uncertainty (i.e., object classification error) without any reduction in overall accuracy or added computational complexity (\cref{sec:protocol}). We also demonstrate how this enables fine-grained introspection via counterfactual predictions by modifying class probabilities directly to produce ``what-if" predictions (\cref{sec:new_expt}).

Second, we analyze the Lyft Level 5 dataset~\cite{HoustonZuidhofEtAl2020} and show that, while it is currently the only public dataset that contains class probabilities, they are overconfident and thus unsuitable for the study of perceptual uncertainty (\cref{sec:lyft_analysis}).

Finally, we present the Perceptual Uncertainty in Prediction (\textbf{\dataname{}}) dataset:
a new challenging, real-world autonomous driving dataset with complex scenes and unfiltered agent class uncertainties. Our dataset better reflects the challenges of the long tail of current state-of-the-art perception systems, thus enabling others to more effectively study robustness to class uncertainty in trajectory forecasting (\cref{sec:pup_analysis}). Our experiments on the Lyft~\cite{HoustonZuidhofEtAl2020} and \dataname{} datasets show that \algname{} significantly improves upon existing approaches, thanks to the inclusion of class probabilities.

\section{RELATED WORK}
{\bf Modular Trajectory Forecasting.} 
Modular methods decompose autonomous driving into distinct sub-tasks, usually perception, prediction, planning and control \cite{schwarting2018planning}. 
A typical interface between perception and forecasting
communicates only the most likely class and state estimate of each object detected by a perception system. As a result, trajectory forecasting methods usually assume their inputs are known with certainty \cite{lefevre2014survey,ma2019trafficpredict,makansi2019overcoming}. In reality, sensors are imperfect and incorrect assumptions of certainty-equivalence in perception---where only the most likely class estimate is passed 
to prediction (but not its uncertainty)---can lead to disastrous outcomes, as in \cref{fig:problem}.

To the best of our knowledge, prior forecasting work has not yet considered the propagation of class uncertainties through modular systems, but there have been many developments. 
For instance, as forecasting is an inherently multi-modal task (especially at intersections), several recent works have proposed multi-modal probabilistic models, trained using exact-likelihood~\cite{rhinehart2018r2p2,chai2019multipath} or variational inference~\cite{SchmerlingLeungEtAl2018,IvanovicSchmerlingEtAl2018,IvanovicPavone2019,SalzmannIvanovicEtAl2020}. 
Generative Adversarial Networks (GANs) \cite{GoodfellowPouget-AbadieEtAl2014} can generate empirical trajectory distributions from sampling multiple predictions~\cite{GuptaJohnsonEtAl2018,roy2019vehicle}. However, analytic distributions are often more useful for gradient-based planning that minimizes collision likelihood~\cite{schwarting2018planning}. Thus, we focus on methods that predict analytic trajectory distributions.

{\bf End-to-End Prediction.} 
End-to-end prediction methods perform detection, tracking, and prediction jointly, operating directly on raw sensor data. FaF \cite{luo2018fast} introduced the approach of projecting LiDAR points into a bird's eye view (BEV) grid, and generating predictions through inferred future detections. 
This approach was extended by IntentNet \cite{casas2018intentnet}, which incorporated HD map information and predicted agent intent. SpAGNN \cite{casas2020spagnn} modeled agent interactions using a graph network, and ILVM \cite{casas2020implicit} extended it by modeling the joint distribution over future trajectories with a latent variable model. These methods only consider homogeneous agents (vehicles); however, MultiXNet \cite{djuric2020multinet} recently extended the BEV approach to heterogeneous agents using separate outputs per agent class. While this approach accounts for class uncertainty, the number of predicted trajectories scales with the number of classes or requires a hard selection of the class 
in the planner.
Broadly, end-to-end methods only incorporate class probabilities implicitly, making it difficult to transparently analyze, probe (e.g., via counterfactual analysis), and understand the effects of perceptual uncertainty. 

{\bf Uncertainty Propagation.}
Methods for propagating uncertainty through neural networks broadly view input data as noisy samples of a true underlying data distribution, and focus on both estimating the true distribution as well as propagating its uncertainty to the output. Towards this end, Bayesian neural networks~\cite{Wright1999,WangShiEtAl2016} and Markov models~\cite{AstudilloNeto2011} are commonly applied. Our work differs as it does not need to perform estimation; object classifiers can fully characterize their output confidence, e.g., as a Categorical distribution over classes, and provide it to downstream modules. 

\section{PROBLEM FORMULATION}
We aim to generate plausible future trajectory distributions for a time-varying number $N(t)$ of diverse interacting agents $A_1,\dots,A_{N(t)}$. Each agent $A_i$ has a class $C_i$ taking one of $K$ values (e.g., Car, Bicycle, Pedestrian). At each time $t$, an upstream perception model estimates the probability that agent $A_i$ is of class $k=1, ..., K$, producing a vector of class probabilities constrained to the $(K-1)$-simplex $\smash{\hat{\mathbf{c}}_i^{(t)} \in \Delta^{K-1}}$ for all agents, where $\hat{c}_{i,k}^{(t)} = p(C_i = k; t)$ is the perception-estimated probability that agent $A_i$ is of class $k$ at time $t$, and
$\Delta^{(K-1)} = \{\mathbf{v} \in \mathbb{R}^K \mid \sum_{k=1}^K v_i = 1 \text{ and } v_i \geq 0\ \forall\ i \}$.
At time $t$, given the state $\mathbf{s}_i^{(t)} \in \mathbb{R}^D$ of each agent (e.g., $x, y$ positions, velocities, and accelerations), their estimated class probabilities $\hat{\mathbf{c}}_i^{(t)}$, and their histories for the previous $H$ timesteps, which we denote as $\mathbf{x} = \mathbf{s}_{1,\dots,N(t)}^{(t - H : t)} \in \mathbb{R}^{(H + 1) \times N(t) \times D}$ and $\smash{\hat{\mathbf{c}} = \hat{\mathbf{c}}_{1,\dots,N(t)}^{(t - H : t)} \in \mathbb{R}^{(H + 1) \times N(t) \times K}}$,
our goal is to produce a distribution over all agents' future states for the next $T$ timesteps, $\mathbf{y} = \mathbf{s}_{1,\dots,N(t)}^{(t + 1 : t + T)} \in \mathbb{R}^{T \times N(t) \times D}$, which we denote as $p(\mathbf{y} \mid \mathbf{x}, \hat{\mathbf{c}})$.

\begin{figure*}[t]
    \centering
    \includegraphics[align=c,width=0.27\textwidth]{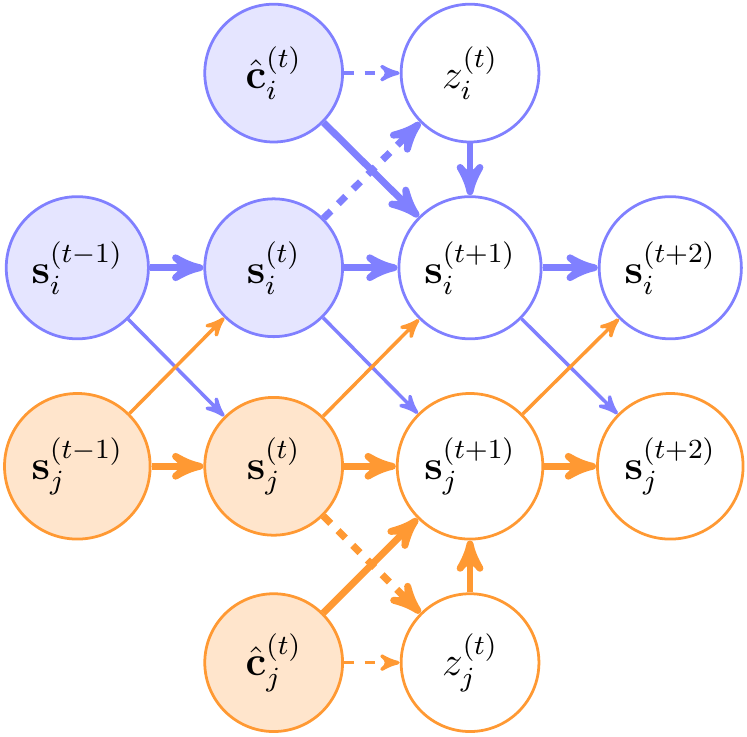}
    \hfill
    \includegraphics[align=c,width=0.67\textwidth]{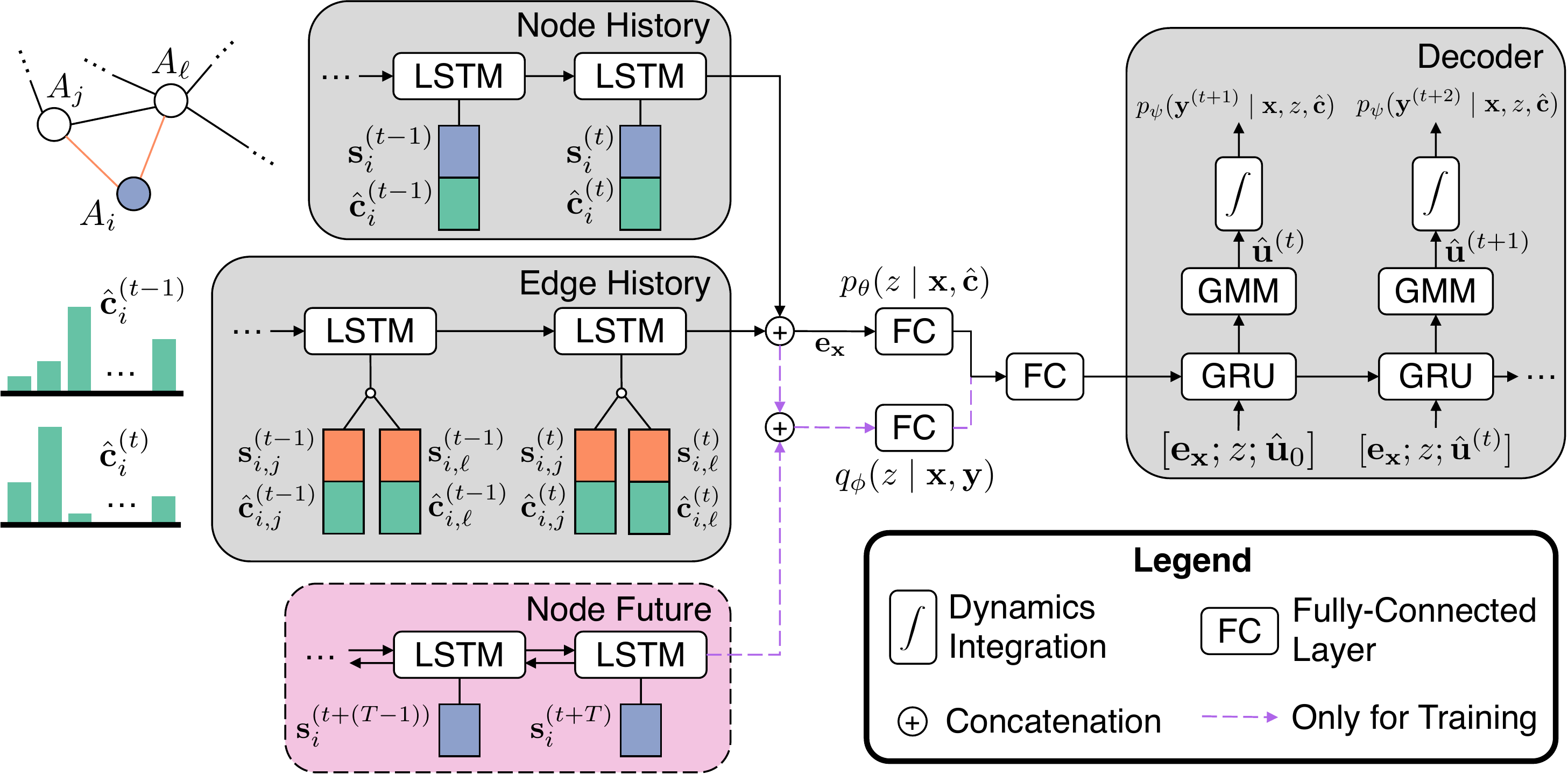}
    \caption{
    \textbf{Left}: Our inference-time probabilistic graphical model for forecasting taking into account agent interactions and class probabilities over time, illustrated for two agents (blue and orange). Known values are shaded and we use thick-arrow notation for carry-forward dependencies~\cite{rhinehart2019precog}.
    Perception provides class-probabilities for each agent $\hat{\mathbf{c}}$, and the known (shaded) past states of both agents help us infer (not generate---denoted by dashed lines) its intentions $z$ at the current time $t$, which help predict the future states of agents.  
    \textbf{Right}: Our approach's network architecture, incorporating agent class uncertainty by encoding class probability values alongside the agent's state.
    }
    \label{fig:architecture}
    
    \vspace*{-0.5cm}
    
\end{figure*}

\section{INCORPORATING CLASS UNCERTAINTY IN TRAJECTORY FORECASTING}\label{sec:method}

We build upon the Trajectron++ \cite{SalzmannIvanovicEtAl2020} framework to implement \algname{}\footnote{All code, models, dataset samples, and dataset links will be made available at \url{https://github.com/TRI-ML/HAICU}.}, due to its ability to perform multi-agent, multi-class trajectory forecasting and public codebase. In this section, we summarize the core components of the algorithm and highlight our key augmentations for incorporating class uncertainty. \cref{fig:architecture} visualizes \algname{}'s probabilistic graphical model and network architecture.

{\bf Input Representation.} We first abstract the scene as an undirected spatiotemporal graph $G = (V, E)$, where nodes represent agents and edges represent their interactions. 
We use the $\ell_2$ distance as a proxy for agent interaction: an undirected edge connects $A_i$ and $A_j$ if $\| \mathbf{p}_i - \mathbf{p}_j \|_2 \leq d$ where $\mathbf{p}_i, \mathbf{p}_j \in \mathbb{R}^2$ are the 2D positions of agents $A_i, A_j$, respectively, and $d$ is a chosen distance threshold. This differs from methods that use a directed graph structure (e.g., Trajectron++), the creation of which relies on hard agent classes to determine edge type and direction.

{\bf Encoding Agent History.} With this graph in hand, our model focuses on encoding a node's state history and how it is influenced by its neighbors. To encode an agent's observed trajectory history, its current and previous states $\mathbf{s}_{1,\dots,N(t)}^{(t - H : t)} \in \mathbb{R}^{(H+1) \times N(t) \times D}$ are fed into a Long Short-Term Memory (LSTM) network \cite{HochreiterSchmidhuber1997} with 32 hidden dimensions. Since we are interested in modeling trajectories, the states $\mathbf{s}_{i}^{(t)}$ are positions, velocities, and accelerations, which are easily estimated online.

{\bf Modeling Agent Interactions.} To model neighboring agents' influence on the modeled agent, edge features from neighboring agents are aggregated via an element-wise sum. We choose to combine features in this way rather than with averaging or concatenation to handle a variable number of neighboring nodes with a fixed architecture while preserving count information \cite{BattagliaPascanuEtAl2016,IvanovicSchmerlingEtAl2018,JainZamirEtAl2016,IvanovicPavone2019,SalzmannIvanovicEtAl2020}. These aggregated states are then fed into an LSTM with 8 hidden dimensions, yielding a single influence representation vector encoding the effect that all neighboring nodes have. The node history and edge influence encodings are then concatenated to produce a single representation vector, $e_\mathbf{x}$.

{\bf Accounting for Multimodality.} Our model leverages the Conditional Variational Autoencoder (CVAE) latent variable framework \cite{SohnLeeEtAl2015} to explicitly account for high-level multimodality in behavior. It produces the target $p(\mathbf{y} \mid \mathbf{x}, \hat{\mathbf{c}})$ distribution by introducing a discrete Categorical latent variable $z \in Z$ which encodes high-level latent behavior and allows for the desired distribution $p(\mathbf{y} \mid \mathbf{x}, \hat{\mathbf{c}})$ to be expressed as
$p(\mathbf{y} \mid \mathbf{x}, \hat{\mathbf{c}}) = \sum_{z \in Z} p_\psi(\mathbf{y} \mid \mathbf{x}, z, \hat{\mathbf{c}}) p_\theta(z \mid \mathbf{x}, \hat{\mathbf{c}})$,
where $|Z| = 25$ and $\psi, \theta$ are network weights. We chose $|Z|$ as such because it allows for the modeling of a wide variety of high-level latent behaviors and any unused latent classes will be ignored by the CVAE \cite{ItkinaIvanovicEtAl2019}.

{\bf Generating Trajectories.} The latent variable $z$ and node representation vector $e_\mathbf{x}$ are then fed into the decoder, a 128-dimensional Gated Recurrent Unit (GRU) \cite{ChoMerrienboerEtAl2014}. Each GRU cell outputs the parameters of a bivariate Gaussian distribution over control actions $\mathbf{u}^{(t)}$ (e.g., velocity). The agent's system dynamics are then integrated with $\mathbf{u}^{(t)}$ to obtain trajectories in position space~\cite{Kalman1960,ThrunBurgardEtAl2005EKF}.
Since the only uncertainty at prediction time stems from \algname{}'s output, and we model agents with linear dynamics, i.e., single integrators, the resulting system dynamics are linear Gaussian. We model all agents as single integrators because we do not know their classes \textit{a priori}. The single integrator model has no constraints, allowing for all possible agent movement. By comparison, e.g., the dynamically-extended unicycle \cite{LaValle2006BetterUnicycle} posits that agents are subject to non-holonomic constraints \cite{PadenCapEtAl2016}, over-constraining pedestrians.

Using one agent dynamics model simplifies \algname{}'s construction as all agents can use the same overall architecture. A different route is to include various agent dynamics, and one way of doing this in \algname{} is to make the decoder multi-headed (i.e., using a different decoder per dynamics model). Such architectures are very popular in the literature, and we explore this avenue in \cref{sec:protocol}.

{\bf Incorporating Class Uncertainty.} To incorporate class probabilities in our model, we concatenate the input class probability vector $\smash{\hat{\mathbf{c}}_i^{(t)}}$ with the state $\smash{\mathbf{s}_i^{(t)}}$ and encode the resulting $(D+K)$-dimensional vector in the same way as the original state vector, with the node and edge history encoders. Neighboring agent class probability vectors are similarly aggregated in the edge encoder.
Concretely, the reason why a CVAE can associate input uncertainty patterns to output trajectories is because the decoder $p_\psi(\mathbf{y} \mid \mathbf{x}, z, \hat{\mathbf{c}})$ directly conditions on the input probabilities $\hat{\mathbf{c}}$.

{\bf Training the Model.} We adopt the same discrete InfoVAE \cite{ZhaoSongEtAl2019} objective function as in Trajectron++. Formally, for each training example $(\{\mathbf{x}_i, \mathbf{\hat{\mathbf{c}}}_i\}, \mathbf{y}_i)$, we aim to maximize
\begin{equation}\label{eqn:loss_fn}
\begin{aligned}
\mathbb{E}&_{z \sim q_\phi(\cdot \mid \mathbf{x}_i, \mathbf{y}_i)} \big[\log p_\psi(\mathbf{y}_i \mid \mathbf{x}_i, z, \hat{\mathbf{c}}_i)\big]\\
&- \beta D_{KL}\big(q_\phi(z \mid \mathbf{x}_i, \mathbf{y}_i) \parallel p_\theta(z \mid \mathbf{x}_i, \hat{\mathbf{c}}_i)\big) + I_{q} (\mathbf{x}_i; z),
\end{aligned}
\end{equation}
where $\phi, \theta, \psi$ are network weights and $I_q$ is the mutual information between $\mathbf{x}_i$ and $z$ under the distribution $q_\phi(\mathbf{x}_i,z)$. To compute $I_q$, we approximate $q_\phi(z \mid \mathbf{x}_i, \mathbf{y}_i)$ with \mbox{$p_\theta(z \mid \mathbf{x}_i)$} and obtain the unconditioned latent distribution by summing out $\mathbf{x}_i$ over the batch \cite{ZhaoSongEtAl2019}.
During training, a bi-directional LSTM with 32 hidden dimensions is used to encode a node’s ground truth future trajectory, producing $q_\phi(z \mid \mathbf{x}, \mathbf{y})$ \cite{SohnLeeEtAl2015}.

\section{DATASETS}
\label{sec:datasets}

\begin{figure}[t]
    \centering
    \includegraphics[width=0.49\linewidth]{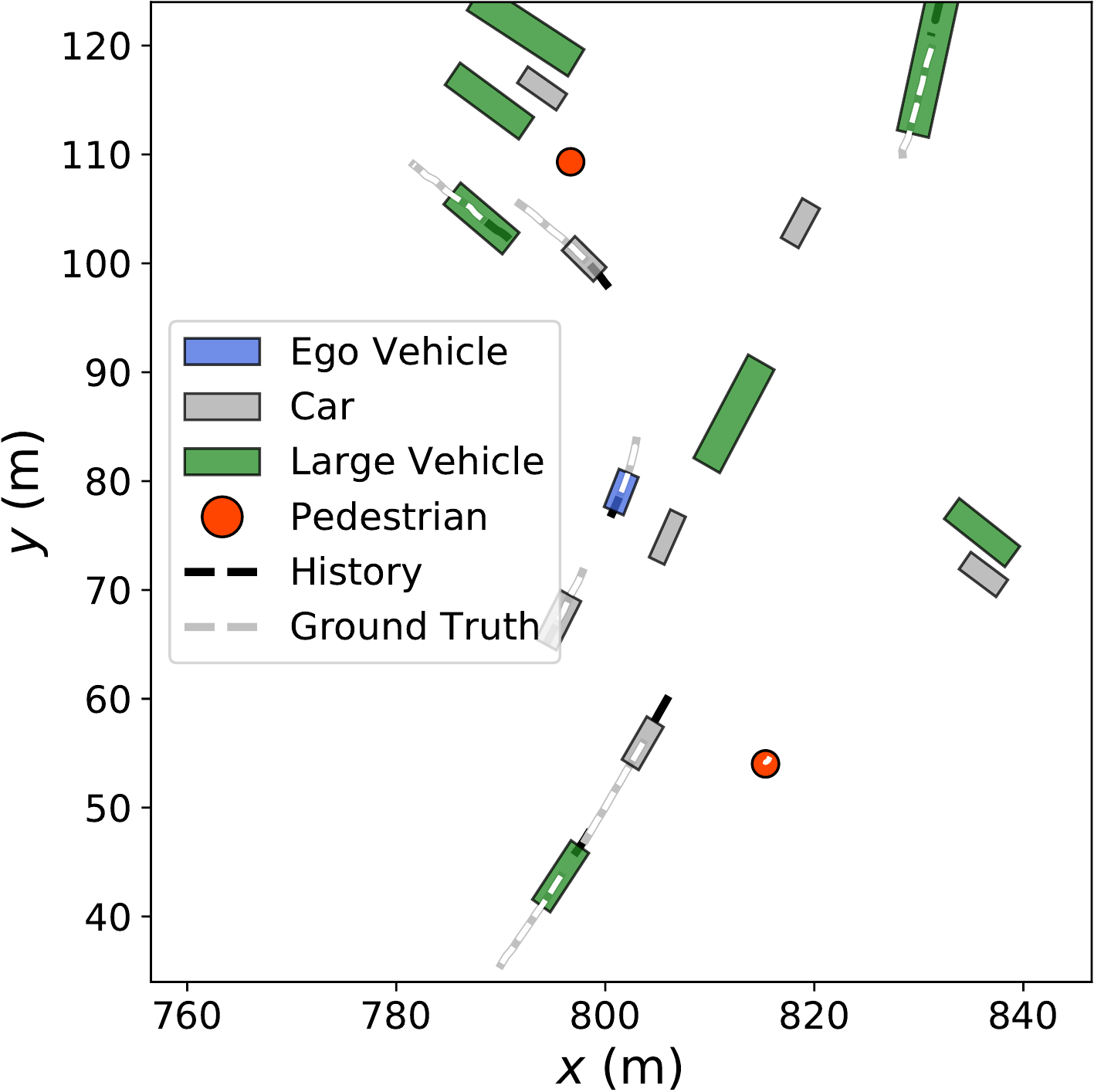}
    \includegraphics[width=0.49\linewidth]{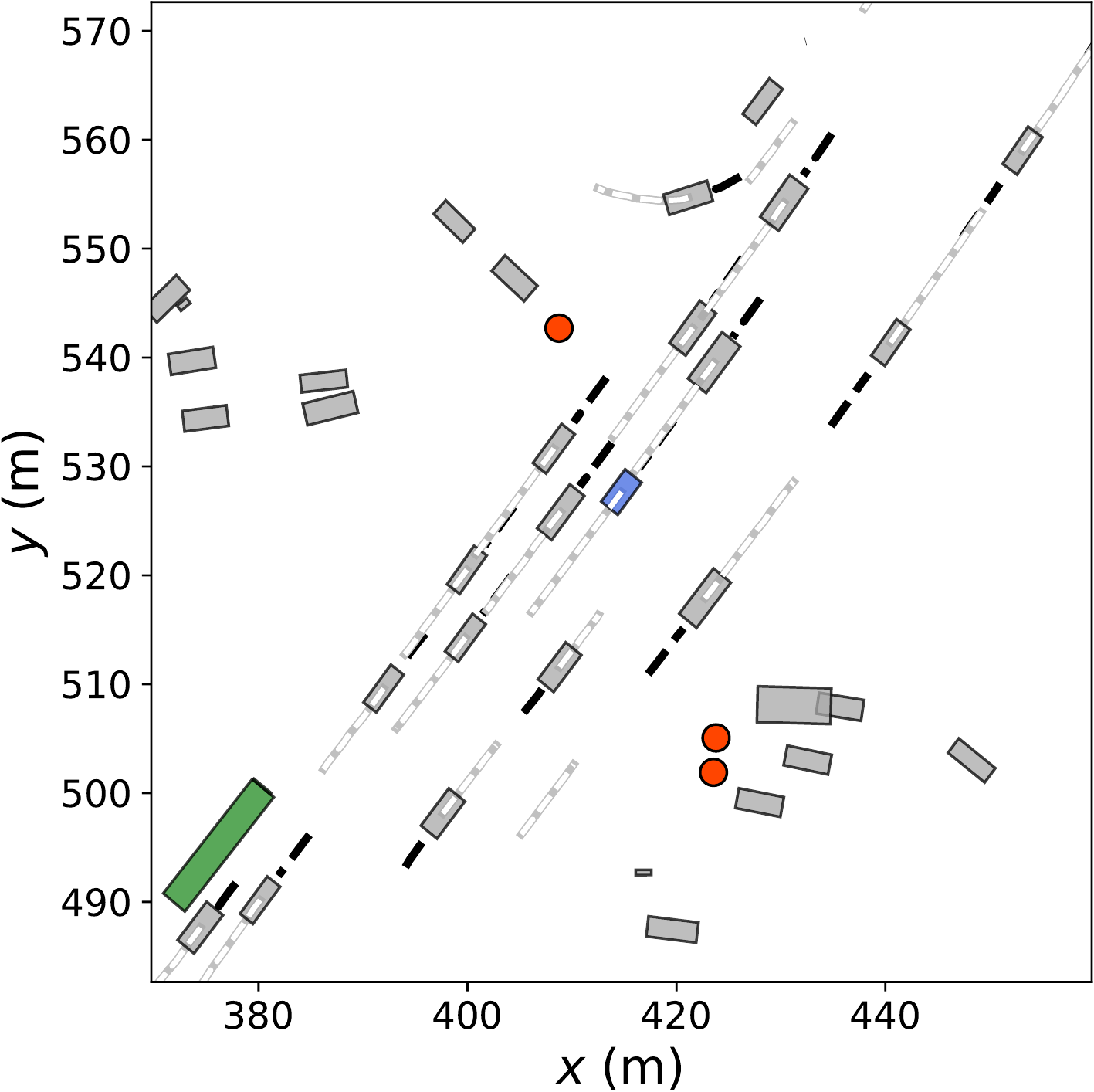}
    \caption{Example scenes from our new \dataname{} dataset.}
    \label{fig:new_example}
    
    \vspace*{-0.35cm}
    
\end{figure}

We evaluate \algname{} on two real-world autonomous driving datasets described in the following sections: Lyft Level 5 \cite{HoustonZuidhofEtAl2020} and \dataname{}, a new dataset that we are releasing with this work. 
Table~\ref{tab:new_lyft_comparison} contains detailed statistics for both datasets and Figure~\ref{fig:new_example} depicts a few scenes from \dataname{}.
More scene visualizations can be found in \cref{supp:more_pup_figs}.

\subsection{Lyft Level 5 Dataset}\label{sec:lyft_analysis}

The Lyft Level 5 dataset is comprised of 1,118 hours of data collected in Palo Alto, USA. Each scene is annotated at 10 Hz ($\Delta t = 0.1s$) and is 25s long, containing 4 agent classes. Importantly, the Lyft dataset was the first, and so far only, to release class probabilities for each detected agent.

{\bf Prevalence of Class Switching.} A key motivation of this work is building trajectory forecasting methods that are robust to perceptual classification errors and uncertainty. In the Lyft Level 5 dataset, we find that $2.1\%$ of all agents experience class-switching, i.e., their highest probability class changes during observation. While the most common switches are between ``unknown" and known classes, there are $14.5$k agents with known class-to-class switches.
For example, \cref{supp:lyft_class_switching} visualizes a scenario where a nearby car is misclassified as a pedestrian in the middle of an intersection. 
Another example in \cref{supp:lyft_class_switching} shows a pedestrian adjacent to the ego-vehicle being misclassified as a car while waiting to cross the street, demonstrating that class switching is not solely due to an agent being very far away from sensor view. 

{\bf Class Switches are Long-lived.} These misclassifications are not solely short-lived, temporary switches, either. We initially applied a temporal smoothing strategy, majority voting with a 5-timestep (0.5s) window, in an attempt to remove cases of high-frequency class switching, but very few switching instances were corrected (only $1$--$3\%$). Further, applying explicit temporal smoothing brings additional tradeoffs regarding accuracy and latency, both of which are especially important for an online task such as object detection and tracking in autonomous driving, but ultimately out of scope for this work.

{\bf Overconfidence.} Even in the presence of class switches, we find the Lyft dataset's class probabilities to be overconfident, i.e., nearly always one-hot vectors, a common issue in deep learning~\cite{GuoPleissEtAl2017}. \cref{tab:new_lyft_comparison} (right) shows the overall average entropy $S_\text{probs}$ of each agent's class probabilities,
    $S_\text{probs} = - \sum_k P(C_i = k) \log P(C_i = k)$,
where $S_\text{probs} = 0.00$ is a distribution with one class having probability $1.0$. 
This certainty is also present over time,
\cref{supp:lyft_prob_ml_class} shows that the most-likely class has more than $90\%$ probability on average.

{\bf Motivation for \dataname{}.} As discussed, the Lyft Level 5 dataset provides a vast amount of data in the regime where perception systems are very certain of their outputs. We argue, however, that perception systems will not always be so certain, and wish to investigate the benefits of incorporating such information in trajectory forecasting. Since we could not find any existing datasets that provide data in this uncertain regime, we present our own.
\begin{table}[t]
\fontsize{9}{9}\selectfont
\centering
\caption{Class counts and uncertainties for \dataname{} compared to Lyft~\cite{HoustonZuidhofEtAl2020}.
}
\begin{tabular}{l|cc|cc}
\toprule
\multicolumn{1}{c|}{} & \multicolumn{2}{c|}{\textbf{\dataname{} (Ours)}} & \multicolumn{2}{c}{\textbf{Lyft Level 5 \cite{HoustonZuidhofEtAl2020}}} \\ \cline{2-5} 
\multicolumn{1}{c|}{\textbf{Class}} & Num. (\%) & $S_\text{probs}$ & Num. (\%) & $S_\text{probs}$ \TBstrut \\ \midrule
bicycle & $1.2$k $(0.8)$ & $1.60$ & $0.1$M $(0.4)$ & $0.09$ \\
car & $117$k $(81.9)$ & $1.10$ & $5.0$M $(24.5)$ & $0.00$ \\
largevehicle & $18$k $(12.4)$ & $1.30$ & $-$ & $-$ \\
motorcycle & $0.5$k $(0.3)$ & $1.57$ & $-$ & $-$ \\
pedestrian & $6.3$k $(4.4)$ & $1.44$ & $0.7$M $(3.3)$ & $0.01$\\
unknown & $0.2$k $(0.1)$ & $0.05$ & $14.6$M $(71.8)$ & $0.00$ \\
\bottomrule
\end{tabular}

\vspace*{-0.15cm}

\label{tab:new_lyft_comparison}
\end{table}

\begin{figure}[t]
    \centering
    \includegraphics[width=\linewidth]{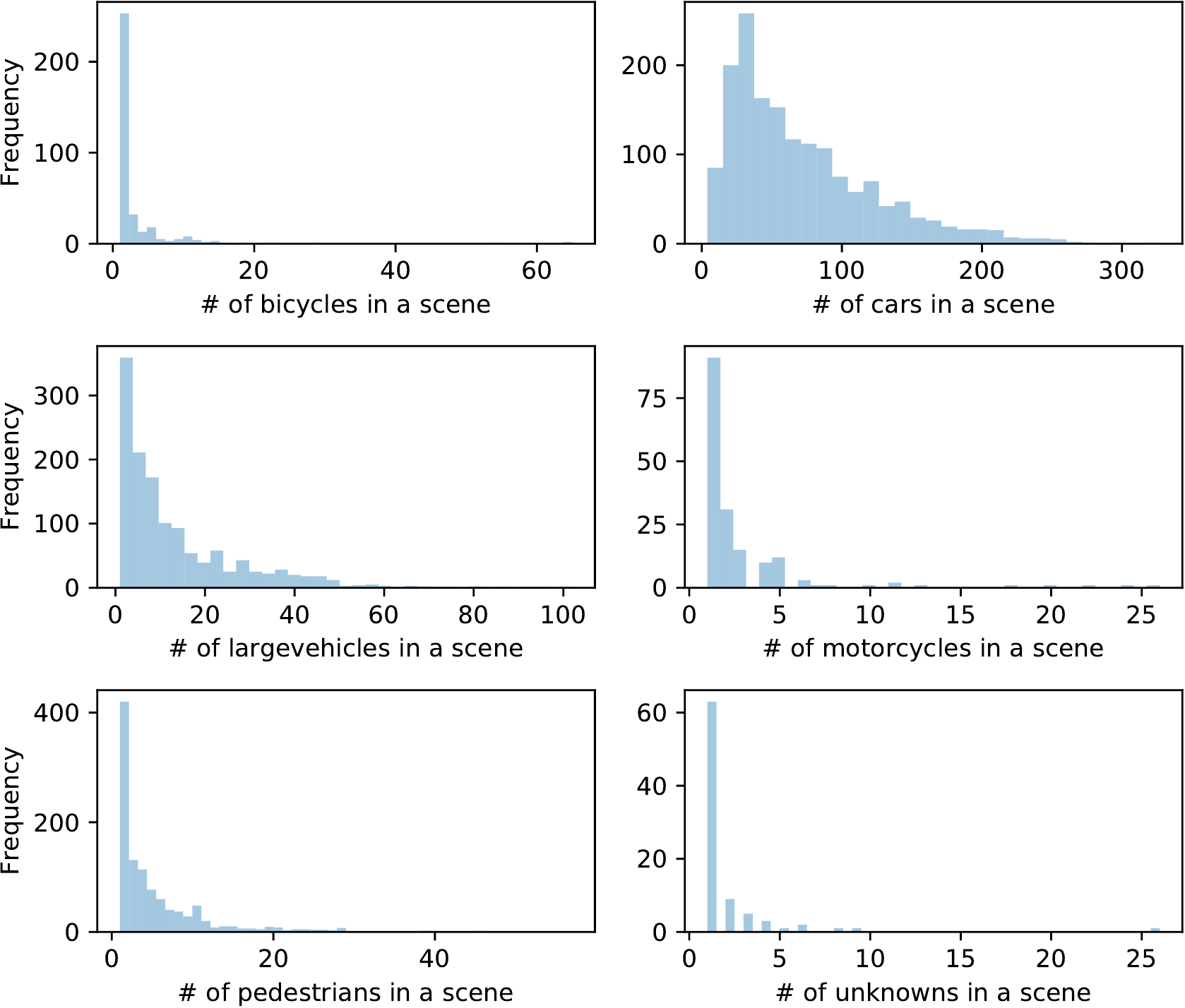}
    \caption{Distribution of the number of agents of a specific type in a scene. For instance, there are more than 200 scenes with 30 cars in them and seldom any scenes with more than 10 unknown objects.}
    \label{fig:pup_scene_histograms}
    
    \vspace*{-0.5cm}
    
\end{figure}

\subsection{\dataname{} Dataset}\label{sec:pup_analysis}

To provide more options for studying the effects of perceptual uncertainty in downstream tasks, one of our core contributions is
\dataname{}, a novel real-world autonomous driving dataset comprised of 1,637 distinct scenes collected with a state-of-the-art self-driving fleet in Tokyo, Japan.
Each scene is 10s long, annotated at 10 Hz with ground-truth trajectories and classes. There are 11 unique agent classes (6 mobile, 4 static, and ``unknown"). Each agent has time-varying class probabilities produced by a state-of-the-art in-house camera- and LiDAR-based perception stack running in production. \cref{fig:pup_scene_histograms} shows the distribution of agents within the \dataname{} dataset's scenes. In particular, it visualizes a histogram over the number of agents of a specific type in a scene, showing that \dataname{} contains scenes with hundreds of unique, diverse agents interacting simultaneously.
Most importantly, since our goal is to quantitatively evaluate the performance of trajectory forecasting in the presence of class uncertainty, we do not perform any post-hoc filtering or smoothing of the perceived agent class probabilities, intentionally releasing the raw frame-by-frame outputs to enable a wide variety of future work (e.g., developing low-latency strategies for temporal smoothing, uncertainty-aware trajectory forecasting, planning under uncertainty).
\cref{tab:new_lyft_comparison} shows a side-by-side comparison of the \dataname{} dataset's class composition and average class uncertainty with those of the Lyft dataset, as measured by mean class probability entropy.
At a high level, \dataname{}'s class probabilities are more uncertain, with far fewer unknown agents.

\begin{figure}[t]
    \centering
    \includegraphics[width=0.8\linewidth]{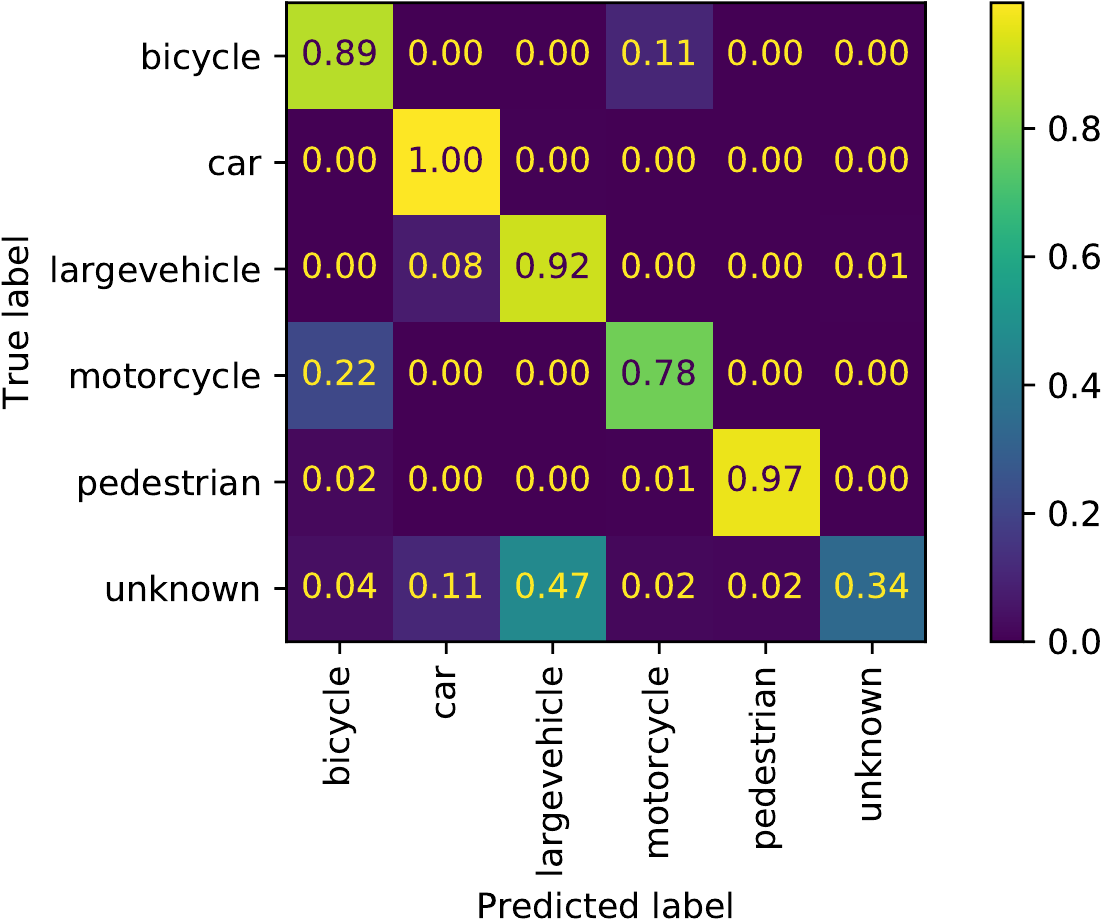}
    
    \vspace*{-0.2cm}
    
    \caption{The (normalized) confusion matrix of the onboard classification system that provides agent class probabilities for the \dataname{} dataset.}
    \label{fig:tri_conf_mat}
    
    \vspace*{-0.5cm}
    
\end{figure}

To quantify the performance of the perception stack used to collect \dataname{}, we evaluate it on a human-annotated dataset from the same region.
\cref{fig:tri_conf_mat} shows the confusion matrix of the object classifier, which is diagonal except for intuitive mistakes, e.g., bicycles and motorcycles.
Accordingly, its top-$k$ accuracies are $96.8\%$, $97.7\%$, $99.2\%$, $99.3\%$, $99.6\%$ for $k=1,\dots,5$, respectively.
For reference, the best top-$1$ classification accuracies on similar object detection tasks are $92$-$97\%$~\cite{HoustonZuidhofEtAl2020,Geiger2012CVPR,CaesarBankitiEtAl2019}. 
Finally, \dataname{} was collected with these AP@0.5 values: $0.75$ for cars, $0.49$ for largevehicles, $0.80$ for pedestrians, and $0.50$ for motorcycles and bicycles.

\begin{table*}[t]
\fontsize{9}{9}\selectfont
\centering
\caption{
Our model significantly outperforms state-of-the-art heterogeneous-agent methods
on the Lyft Level 5 dataset \cite{HoustonZuidhofEtAl2020}. It is expected that our method performs similarly to One-Hot because the Lyft data mostly contains one-hot class probabilities.
Results with the minADE and minFDE metrics can be found in \cref{supp:minADE_minFDE}.
SE = Standard Error.
}

\vspace*{-0.2cm}

\begin{tabular}{l|c|ccc|c|ccc}
\toprule
\multicolumn{1}{c|}{\textbf{Lyft Level 5} \cite{HoustonZuidhofEtAl2020}} & \multicolumn{1}{c}{\textbf{ADE {\scriptsize{$\pm$ SE}}}} & \multicolumn{3}{c|}{\textbf{FDE {\scriptsize{$\pm$ SE}} (m)}} & \multicolumn{1}{c}{\textbf{ANLL {\scriptsize{$\pm$ SE}}}} & \multicolumn{3}{c}{\textbf{FNLL {\scriptsize{$\pm$ SE}} (nats)}} \\ \cline{1-9} 
\multicolumn{1}{c|}{\textbf{Pred. Horizon}} & 3s & 1s      & 2s        & 3s & 3s & 1s      & 2s        & 3s\Tstrut{} \\ \midrule
MATS \cite{IvanovicElhafsiEtAl2020} & $1.21$\scriptsize{$\pm 0.13$} & $0.34$\scriptsize{$\pm 0.04$} & $1.22$\scriptsize{$\pm 0.15$} & $2.90$\scriptsize{$\pm 0.32$} & $3.78$\scriptsize{$\pm 0.34$} & $0.25$\scriptsize{$\pm 0.18$} & $2.47$\scriptsize{$\pm 0.24$} & $12.06$\scriptsize{$\pm 1.01$}\\
Trajectron++ \cite{SalzmannIvanovicEtAl2020} & $0.47$\scriptsize{$\pm 5$e-$3$} & $0.26$\scriptsize{$\pm 3$e-$3$} & $0.60$\scriptsize{$\pm 7$e-$3$} & $1.05$\scriptsize{$\pm 0.01$} & $-1.06$\scriptsize{$\pm 0.02$} & $-1.34$\scriptsize{$\pm 0.02$} & $-0.13$\scriptsize{$\pm 0.02$} & $0.68$\scriptsize{$\pm 0.02$}\\
Multi-Head~\cite{djuric2020multinet} & $0.45$\scriptsize{$\pm 5$e-$3$} & $0.26$\scriptsize{$\pm 3$e-$3$} & $0.58$\scriptsize{$\pm 7$e-$3$} & $0.99$\scriptsize{$\pm 0.01$} & $-1.15$\scriptsize{$\pm 0.02$} & $-1.44$\scriptsize{$\pm 0.02$} & $-0.25$\scriptsize{$\pm 0.02$} & $0.54$\scriptsize{$\pm 0.02$} \\
One-Hot & $0.44$\scriptsize{$\pm 5$e-$3$} & $0.25$\scriptsize{$\pm 2$e-$3$} & $0.56$\scriptsize{$\pm 6$e-$3$} & $0.95$\scriptsize{$\pm 0.01$} & $\mathbf{-1.36}$\scriptsize{$\pm 0.02$} & $-1.56$\scriptsize{$\pm 0.02$} & $\mathbf{-0.47}$\scriptsize{$\pm 0.02$} & $\mathbf{0.27}$\scriptsize{$\pm 0.03$}\\
\hline
\algname{} (ours) & $\mathbf{0.43}$\scriptsize{$\pm 5$e-$3$} & $\mathbf{0.24}$\scriptsize{$\pm 2$e-$3$} & $\mathbf{0.54}$\scriptsize{$\pm 6$e-$3$} & $\mathbf{0.94}$\scriptsize{$\pm 0.01$} & $-1.35$\scriptsize{$\pm 0.02$} & $\mathbf{-1.57}$\scriptsize{$\pm 0.02$} & $-0.45$\scriptsize{$\pm 0.02$} & $0.31$\scriptsize{$\pm 0.03$} \Tstrut{} \\
\bottomrule
\end{tabular}

\label{tab:lyft_quant}

\vspace*{-0.25cm}

\end{table*}

\section{EXPERIMENTS}\label{sec:protocol}

{\bf Baselines.} We compare \algname{} against the following state-of-the-art approaches that also produce multimodal predictions for varying numbers of diverse agents:

(1)~MATS \cite{IvanovicElhafsiEtAl2020}: each scene is modeled with a mixture of affine dynamical systems,
forward-integrated to produce predictions,

(2)~Trajectron++ \cite{SalzmannIvanovicEtAl2020}: a state-of-the-art LSTM-CVAE encoder-decoder whose architecture is based on the spatiotemporal structure of the scene. 

Note that both MATS and Trajectron++ assume perfect agent classification, and rely on such information in their network components (e.g., sharing weights among same-class components).

(3)~Multi-Head: Rather than encoding agent classes in $e_\mathbf{x}$, methods like MultiXNet~\cite{djuric2020multinet} are multi-headed and produce an output for each possible agent type. We implement the same, additionally augmenting each head with a dynamics model (i.e., dynamically-extended unicycle \cite{LaValle2006BetterUnicycle} for vehicles and single integrator for others, as in~\cite{SalzmannIvanovicEtAl2020}); these outputs are then combined in a mixture model where the class probabilities $\hat{\mathbf{c}}_i^{(t)}$ are the mixing probabilities. %

(4)~One-Hot: An ablation of \algname{} with one-hot class probabilities passed in.
This corresponds to the hard class-conditioning of Trajectron++ (no uncertainty modeling) while using our class-agnostic weight-sharing scheme.
{\bf Metrics.} 
We evaluate our approach with a variety of deterministic and probabilistic metrics:
\emph{Average/Final Displacement Error (ADE/FDE)}: mean/final $\ell_2$ distance between the ground truth and predicted trajectories, 
\emph{Average/Final Negative Log-Likelihood (ANLL/FNLL)}: the mean/final NLL of the ground truth trajectory under the predicted distribution. 
\emph{minADE/minFDE}: ADE/FDE between the ground truth and best of 20 samples~\cite{GuptaJohnsonEtAl2018}.

For ADE/FDE we compare methods' single most-likely trajectory prediction (establishing accuracy for the deterministic use case), while for ANLL/FNLL we compute likelihoods using their full output distributions (determining performance for probabilistic use cases). For minADE/minFDE, we randomly sample 20 trajectories from each model and compute the ADE/FDE of the best \cite{GuptaJohnsonEtAl2018}.

{\bf Evaluation Methodology.} For both datasets, we use $70\%, 15\%, 15\%$ data (scene) splits for training, validation, and testing, respectively. Models are trained to predict forward 20 timesteps (2s) from at most 20 timesteps (2s) of observed data. We trained each model until their validation performance stopped improving. 
Further training details
can be found in \cref{supp:training_info}.

\subsection{Lyft Dataset Results}
\label{sec:lyft_expt}
\cref{tab:lyft_quant} summarizes our evaluation on the Lyft dataset, and shows that \algname{} outperforms state-of-the-art trajectory forecasting methods on the probabilistic ANLL and FNLL metrics, and is competitive on the deterministic ADE and FDE metrics. Notably, even though the Lyft dataset does not have much class uncertainty (\cref{tab:new_lyft_comparison}), our method and its ablations still outperform MATS and Trajectron++, which architecturally incorporate agent class information, indicating that a class-agnostic modeling scheme yields improvements. This is also the reason why our method with one-hot class probabilities performs similarly to our method with full probability input, as the Lyft dataset is already mostly comprised of one-hot class probability vectors (\cref{tab:new_lyft_comparison}). 

Further, encoding class probabilities with the state input outperforms a multi-headed output mixture. This is likely due to the multi-headed version of our model being an interpolation between Trajectron++ (separate encoder and decoder components per class) and \algname{} (same encoder and decoder components for all classes), using the same encoder for all classes but class-specific decoder.

Finally, while all models were trained with a prediction horizon of $2s$, we also evaluate their performance on a $3s$ prediction horizon as an additional test of temporal generalization. As can be seen in \cref{tab:lyft_quant}, \algname{} maintains strong performance at longer time horizons. 
Additional results per agent class can be found in \cref{supp:additional_lyft_results}.

\subsection{\dataname{} Dataset Results}
\label{sec:new_expt}

\begin{table*}[t]
\fontsize{9}{9}\selectfont
\centering
\caption{
Our model significantly outperforms existing methods
on our new \dataname{} dataset.
Results with the minADE and minFDE metrics reinforce this and can be found in \cref{supp:minADE_minFDE}.
SE = Standard Error.
}
\begin{tabular}{l|c|ccc|c|ccc}
\toprule
\multicolumn{1}{c|}{\textbf{PUP}} & \multicolumn{1}{c}{\textbf{ADE {\scriptsize{$\pm$ SE}}}} & \multicolumn{3}{c|}{\textbf{FDE {\scriptsize{$\pm$ SE}} (m)}} & \multicolumn{1}{c}{\textbf{ANLL {\scriptsize{$\pm$ SE}}}} & \multicolumn{3}{c}{\textbf{FNLL {\scriptsize{$\pm$ SE}} (nats)}} \\ \cline{1-9} 
\multicolumn{1}{c|}{\textbf{Pred. Horizon}}         & 3s & 1s      & 2s        & 3s & 3s & 1s      & 2s        & 3s\Tstrut{} \\ \midrule
MATS \cite{IvanovicElhafsiEtAl2020} & $1.23$\scriptsize{$\pm 0.21$} & $0.48$\scriptsize{$\pm 0.12$} & $1.08$\scriptsize{$\pm 0.19$} & $2.12$\scriptsize{$\pm 0.35$} & $5.66$\scriptsize{$\pm 0.57$} & $2.23$\scriptsize{$\pm 0.65$} & $3.59$\scriptsize{$\pm 0.54$} & $11.15$\scriptsize{$\pm 1.02$}\\
Trajectron++ \cite{SalzmannIvanovicEtAl2020} & $0.75$\scriptsize{$\pm 0.02$} & $0.48$\scriptsize{$\pm 0.01$} & $0.93$\scriptsize{$\pm 0.02$} & $1.52$\scriptsize{$\pm 0.03$} & $0.04$\scriptsize{$\pm 0.05$} & $-0.30$\scriptsize{$\pm 0.04$} & $0.69$\scriptsize{$\pm 0.07$} & $1.32$\scriptsize{$\pm 0.06$}\\
Multi-Head~\cite{djuric2020multinet} & $0.84$\scriptsize{$\pm 0.06$} & $0.55$\scriptsize{$\pm 0.04$} & $1.05$\scriptsize{$\pm 0.08$} & $1.64$\scriptsize{$\pm 0.16$} & $0.18$\scriptsize{$\pm 0.09$} & $-0.28$\scriptsize{$\pm 0.10$} & $0.86$\scriptsize{$\pm 0.13$} & $1.57$\scriptsize{$\pm 0.15$} \\
One-Hot & $0.69$\scriptsize{$\pm 0.01$} & $0.42$\scriptsize{$\pm 9$e-$3$} & $0.85$\scriptsize{$\pm 0.02$} & $1.41$\scriptsize{$\pm 0.03$} & $-0.23$\scriptsize{$\pm 0.06$} & $-0.59$\scriptsize{$\pm 0.07$} & $0.44$\scriptsize{$\pm 0.07$} & $1.08$\scriptsize{$\pm 0.09$}\\
\hline
\algname{} (ours) & $\mathbf{0.65}$\scriptsize{$\pm 0.02$} & $\mathbf{0.40}$\scriptsize{$\pm 0.01$} & $\mathbf{0.79}$\scriptsize{$\pm 0.02$} & $\mathbf{1.35}$\scriptsize{$\pm 0.04$} & $\mathbf{-0.35}$\scriptsize{$\pm 0.06$} & $\mathbf{-0.77}$\scriptsize{$\pm 0.08$} & $\mathbf{0.32}$\scriptsize{$\pm 0.07$} & $\mathbf{0.96}$\scriptsize{$\pm 0.11$} \Tstrut{}\\
\bottomrule
\end{tabular}

\label{tab:new_quant}

\vspace*{-0.25cm}

\end{table*}

{\bf Quantitative Results.} \cref{tab:new_quant} summarizes the evaluation on the \dataname{} dataset, and shows that \algname{} significantly outperforms state-of-the-art trajectory forecasting methods on both deterministic (two-tailed $t$-test; $P < 0.025$) and probabilistic (two-tailed $t$-test; $P < 10^{-10}$) metrics across all prediction horizons. 
Evaluating with minADE/minFDE reinforces this, and is shown in \cref{supp:minADE_minFDE}.

Notably, in the presence of increased class uncertainty our method now outperforms the one-hot ablated version, significantly so on ANLL (two-tailed $t$-test; $P < 0.02$), verifying that our model is able to effectively use the full input probability information. The same is true for \algname{}'s Top 2 ablation, showing that just adding one more class yields significant benefits over the one-hot ablation. Unlike the Lyft dataset, the multi-headed version of our model does not perform as well. We believe this is a direct result of the increased class uncertainty as the decoder heads are trained together in an overall mixture model. As a result, class uncertainty directly competes with class-based dynamics (the output heads are simultaneously trying to model the same target 
trajectory), leading to a reduction in overall output diversity.

Further, as in \cref{sec:lyft_expt}, all models used a prediction horizon of $2s$ during training and we also evaluate their performance on a longer, $3s$ prediction horizon. As can be seen in \cref{tab:new_quant}, our approach maintains its strong performance over longer time horizons. 
Additional detailed results per agent class can be found in \cref{supp:additional_new_results}.

\begin{figure}[t]
    \centering
    \includegraphics[width=\linewidth]{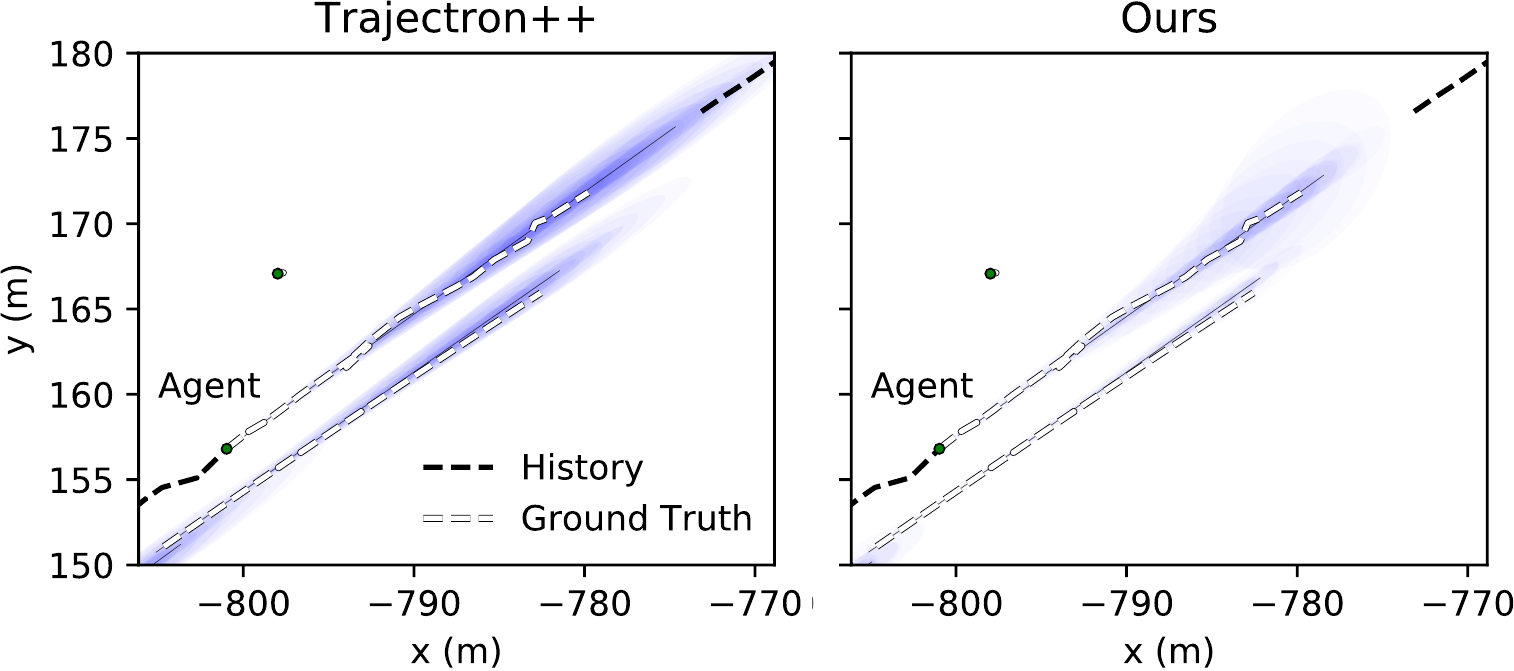}
    \caption{Our method effectively propagates class probability uncertainty through to its outputs. In this example, the marked agent is a vehicle with high class uncertainty (class probability entropy $S_\text{probs} = 2.06$, maximum possible entropy is $\ln(11) \approx 2.40$). Unlike Trajectron++ \cite{SalzmannIvanovicEtAl2020} (\textbf{left}),
    our method is able to incorporate such information and produces much more accurate predictions (\textbf{right}).
    }
    \label{fig:new_qual}
    
    \vspace*{-0.5cm}
    
\end{figure}

\begin{figure*}[t]
    \centering
    \includegraphics[width=\linewidth]{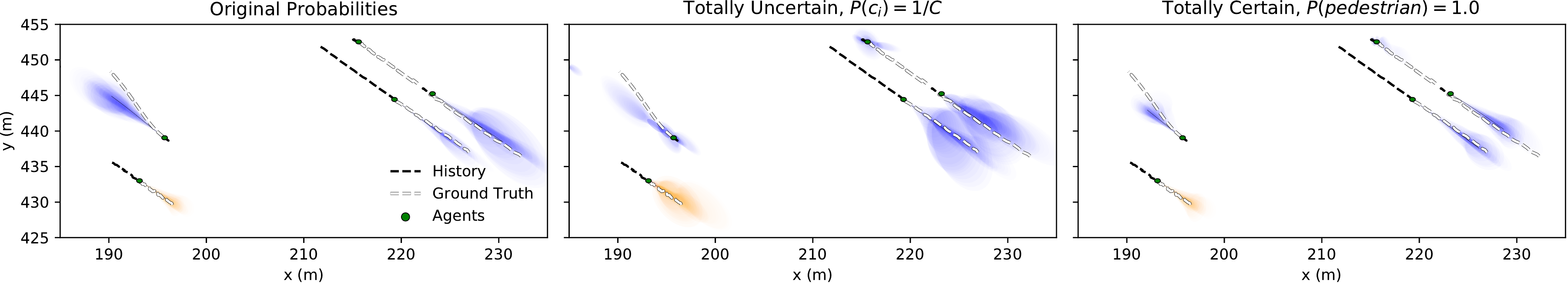}
    
    \vspace*{-0.25cm}
    
    \caption{Our method is able to make counterfactual predictions, i.e., predictions where the input probability distribution is manually modified to produce ``what-if" predictions. Color denotes the original class of the agent (orange for pedestrians, blue for vehicles). \textbf{Left}: Our model's predictions with original class probabilities. \textbf{Middle}: All agents have fully-uncertain class probabilities. \textbf{Right}: All agents have fully-certain pedestrian class probabilities.}
    \label{fig:new_sens_analysis}
\end{figure*}

\begin{figure*}[t]
    \centering
    \includegraphics[width=\linewidth]{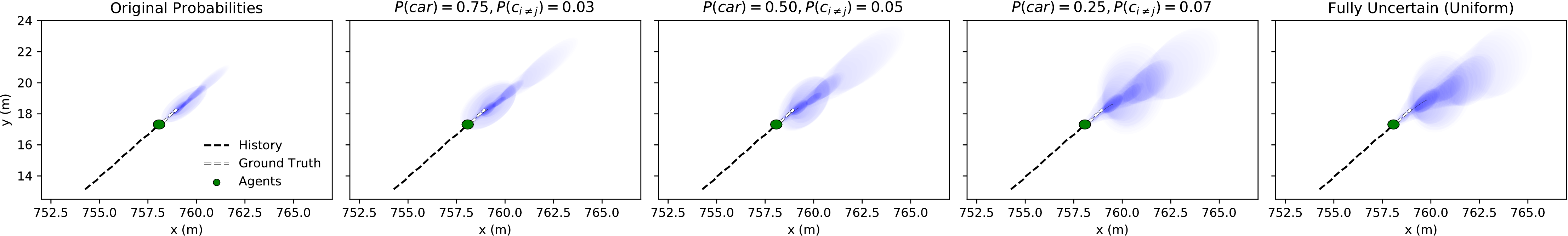}
    
    \vspace*{-0.25cm}
    
    \caption{\algname{}'s counterfactual predictions interpolate smoothly across input probabilities.}
    \label{fig:new_sens_slider}

    \vspace*{-0.5cm}

\end{figure*}

{\bf Qualitative Results.} Incorporating class uncertainty in trajectory forecasting yields qualitative differences in output behavior. \cref{fig:new_qual} visualizes a scene from the \dataname{} dataset with two vehicles moving parallel to each other, whose future trajectories are forecasted by Trajectron++, our one-hot ablated model,
and our model.
In this example, we see that Trajectron++ makes overconfident predictions which overshoot the marked agent's ground truth future trajectory. Our one-hot ablated model version similarly overshoots the ground truth, although with a bit more uncertainty on the left side of the ground truth. In comparison, our method not only significantly more accurately predicts the ground truth, it also produces equal amounts of uncertainty on either side of the ground truth. Specifically, incorporating full probabilities with our approach in this example yielded an ADE and FDE that are $1.9$ m and $3.9$ m less, respectively, than both Trajectron++ and our one-hot ablated model. The predicted distribution is also more accurate, with ANLL and FNLL values that are $0.24$ nats and $0.83$ nats less, respectively, than both Trajectron++ and our one-hot ablated model.

{\bf Counterfactual Predictions.} By explicitly conditioning on class probabilities at the input, our method is additionally able to make \emph{counterfactual} predictions, i.e., predictions where the input class probabilities are manually specified ahead of time in order to produce ``what-if" predictions. In the example visualized in \cref{fig:new_sens_analysis}, a pedestrian is walking towards the bottom-right, a stopped car is starting to move on the left, and two vehicles are driving straight towards the bottom-right. 

Our method's predictions with the original probabilities in the data are shown in \cref{fig:new_sens_analysis} (left). In particular, our model predicts that each agent will mostly keep moving along the same heading with some uncertainty. In \cref{fig:new_sens_analysis} (middle), we manually change the class probabilities for each agent to be a uniform distribution over all classes (representing total class uncertainty). Immediately, we see that our model produces more uncertainty.
Conversely, in \cref{fig:new_sens_analysis} (right) we made the class probabilities totally certain (one-hot for ``pedestrian") for each agent. With this information, our model predicts forward motion at normal human walking speeds for all agents, as desired. This leaves the original bottom-left pedestrian prediction virtually unchanged, but greatly alters the car predictions on the right to match how a pedestrian would behave.

{\bf Prediction Smoothness.} Our model's predictions interpolate smoothly in the space of input probabilities. In \cref{fig:new_sens_slider}, we manually vary the class probabilities of the visualized agent from its original values (high probability of being a car) to fully uncertain values (uniform probability for all classes). We can see that as class uncertainty increases, so does our model's output uncertainty. Prediction smoothness is desirable as it means that our model can smoothly propagate input uncertainty through to its outputs across a wide range of class probability values.

\begin{figure*}[t]
    \centering
    \includegraphics[width=0.95\linewidth]{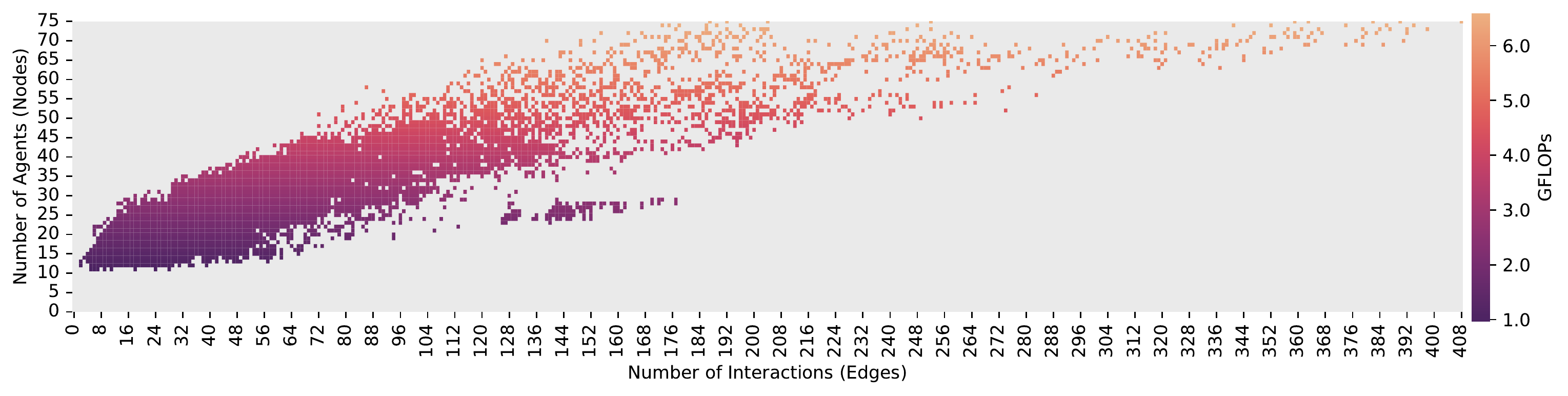}
    \includegraphics[width=0.95\linewidth]{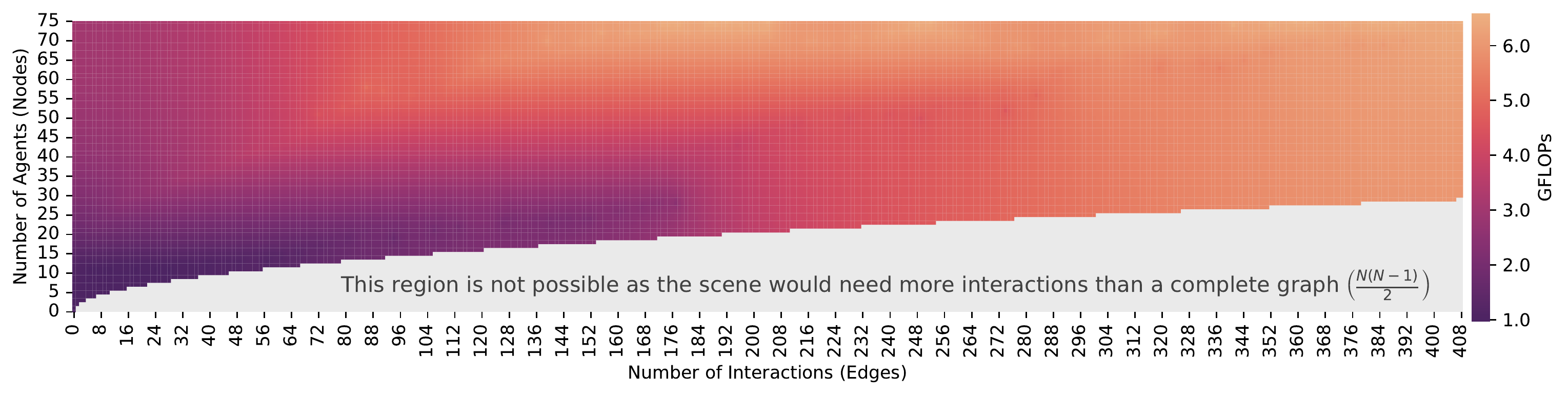}
    
    \vspace*{-0.25cm}
    
    \caption{\textbf{Top:} FLOPs required to run \algname{} on the problem sizes found in the \dataname{} dataset. \textbf{Bottom:} The same, but with an additional optimization-based extrapolation scheme (Laplacian smoothing~\cite{HallacLeskovecEtAl2015}) to impute values for configurations that were not encountered in the \dataname{} dataset.}
    \label{fig:flops}
    
    \vspace*{-0.5cm}
    
\end{figure*}

{\bf Runtime.} 
On the busiest scene (with 75 agents), \algname{} only requires $6.58$ GFLOPs to predict all agents’ futures (for reference, this is $\sim 1$ GFLOP less than executing a forward pass of ResNet-34)\footnote{See \url{https://github.com/Lyken17/pytorch-OpCounter} and \url{https://github.com/sovrasov/flops-counter.pytorch} for other models' FLOP counts (please note that $1$ Multiply-Accumulate (MAC) $= 2$ FLOPs).}. \cref{fig:flops} shows heatmaps visualizing the FLOPs required to run \algname{} on the various scene sizes (in terms of node and edge counts) encountered in the \dataname{} dataset.
Further, due to our aggressive weight-sharing scheme, \algname{} only has $117,389$ parameters.

\section{CONCLUSION}

We investigate the problem of robustness to perceptual uncertainty in multi-agent trajectory forecasting. In particular, we highlight the importance of leveraging the full distribution over the semantic classes of agents which is typically provided by perception models, but often quantized to the (potentially-overconfident) mode.
We introduce a new method (\emphalgname{}) for heterogeneous-agent trajectory forecasting that explicitly incorporates class probabilities, as well as a new autonomous driving dataset (\emphdataname{}) to study the impact of Perceptual Uncertainty in Prediction.
In addition to a more informed representation of uncertainty, our approach also enables new capabilities such as counterfactual predictions, opening up interesting future research in causal reasoning and interpretability for prediction and planning.

\bibliographystyle{IEEEtran}
\bibliography{ASL_papers,main,added}

\clearpage
\newpage

\appendix

\subsection{Additional PUP Scene Visualizations}
\label{supp:more_pup_figs}

\begin{figure}[t]
    \centering
    \includegraphics[width=0.49\linewidth]{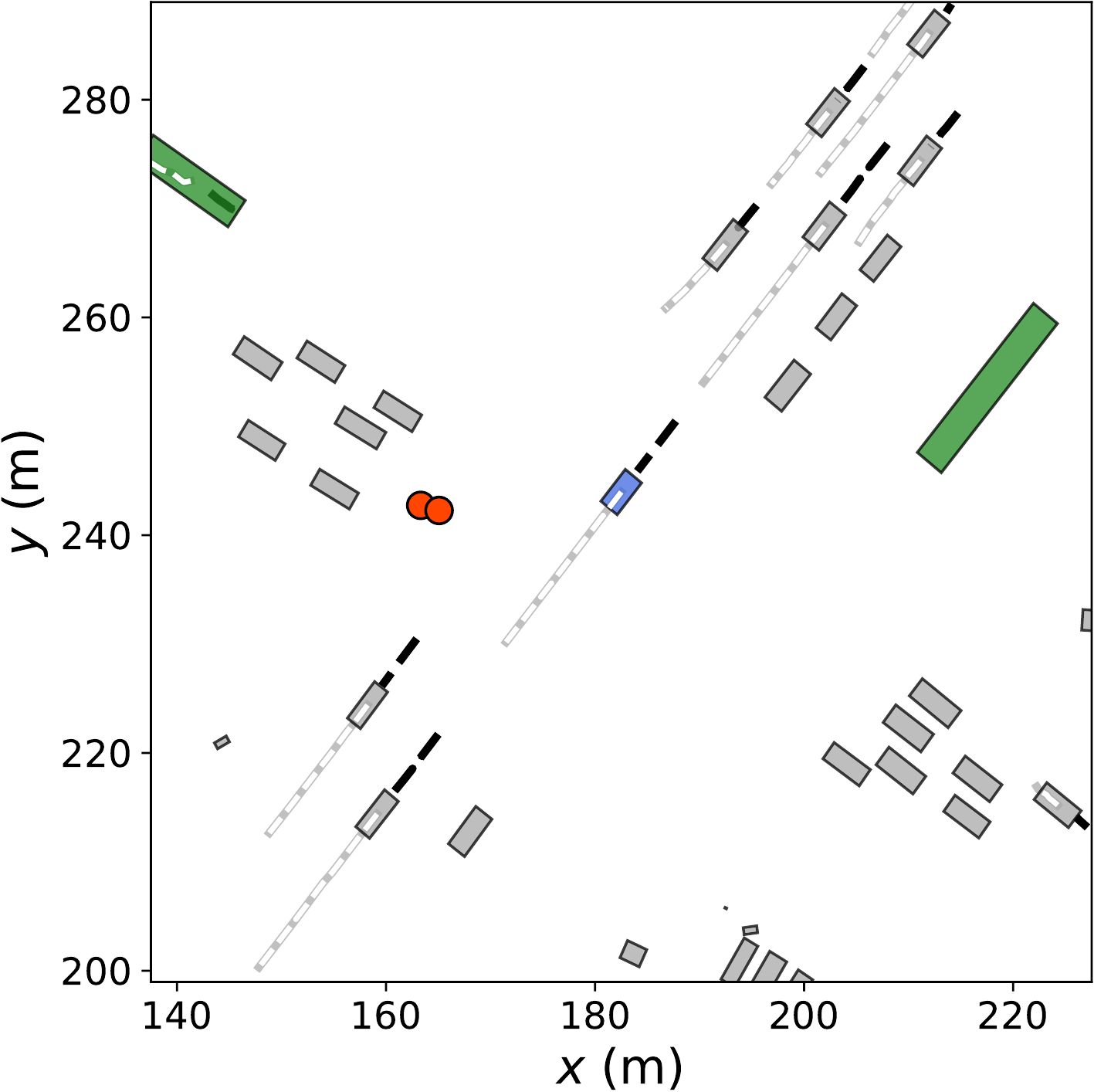}
    \includegraphics[width=0.49\linewidth]{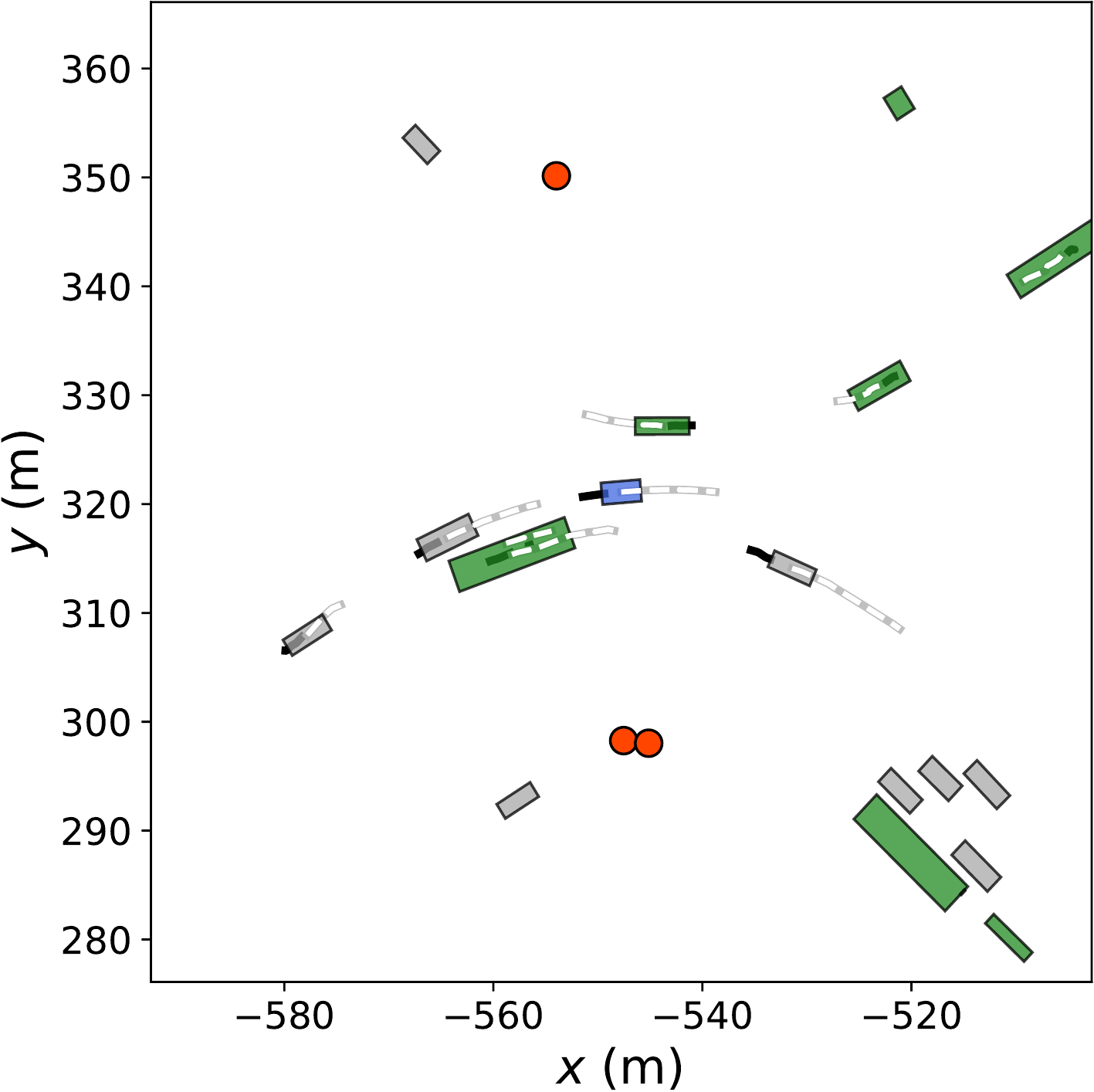}
    \caption{Additional example scenes from our new \dataname{} dataset.}
    \label{fig:supp_new_examples}
\end{figure}

Additional scenes from our \dataname{} dataset are visualized in \cref{fig:supp_new_examples}.

\subsection{Detailed Lyft Dataset Statistics}
\label{supp:detailed_lyft_statistics}

\subsubsection{Class Switching Histogram} 
\begin{figure}[t]
    \centering
    \includegraphics[width=0.9\linewidth]{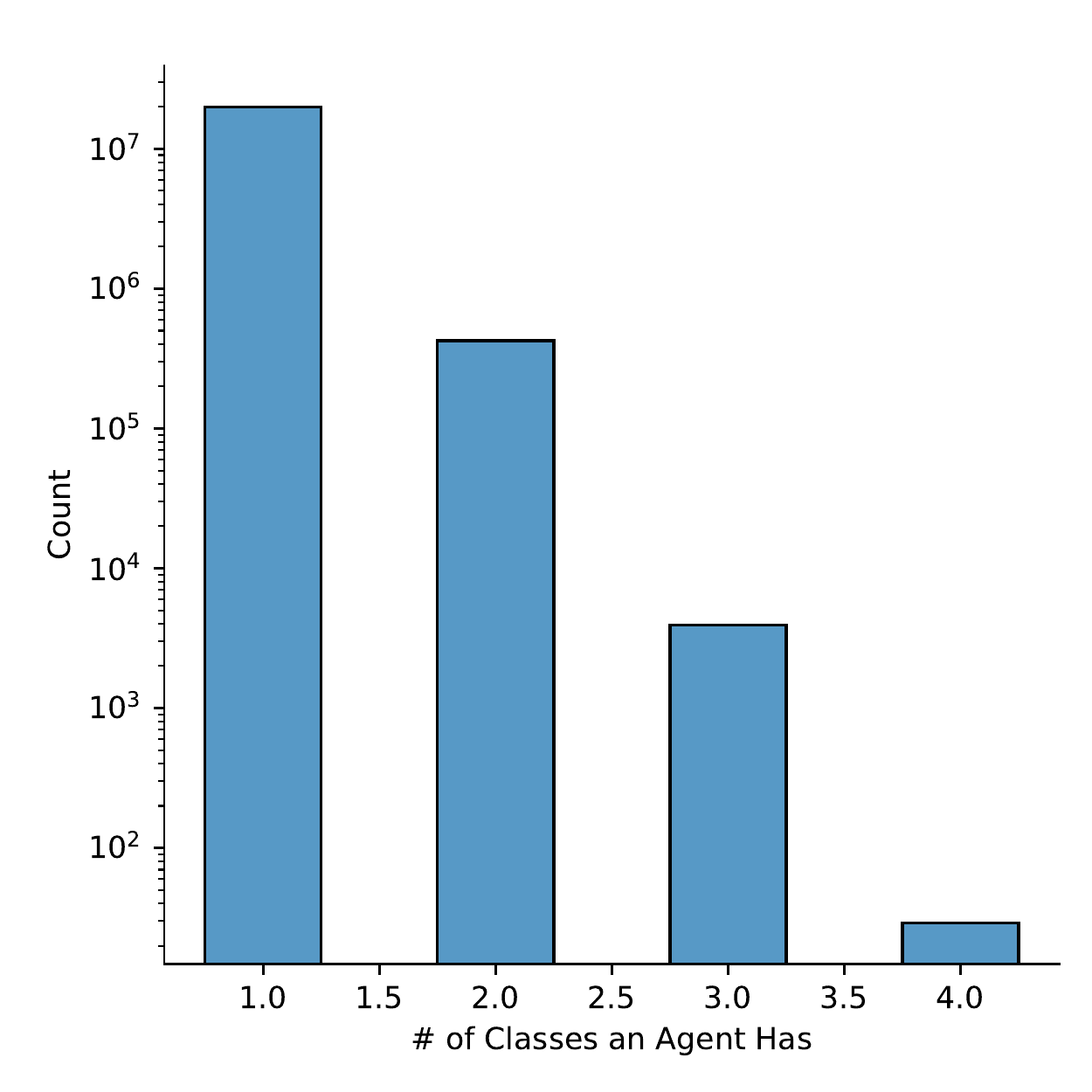}
    \caption{A histogram of the number of most-likely classes an agent has. For example, if an agent has 4 classes then it experiences at least three class switches between 4 distinct classes during observation}
    \label{fig:supp_class_switches_count}
\end{figure}

As mentioned in the main text, we find that $2.1\%$ of all agents in the Lyft Level 5 dataset~\cite{HoustonZuidhofEtAl2020} experience class-switching, i.e., their highest probability class changes at least once during observation. \cref{fig:supp_class_switches_count} dives deeper and shows the number of most-likely classes an agent has. For example, if an agent has 1 class then it experiences no class switches and has a consistent most-likely class throughout observation. Accordingly, if an agent has 2 classes then it experiences at least one class switch between two distinct classes during observation.

\subsubsection{Examples of Class Switching}\label{supp:lyft_class_switching}
\begin{figure*}[t]
    \centering
    \includegraphics[width=0.7\linewidth]{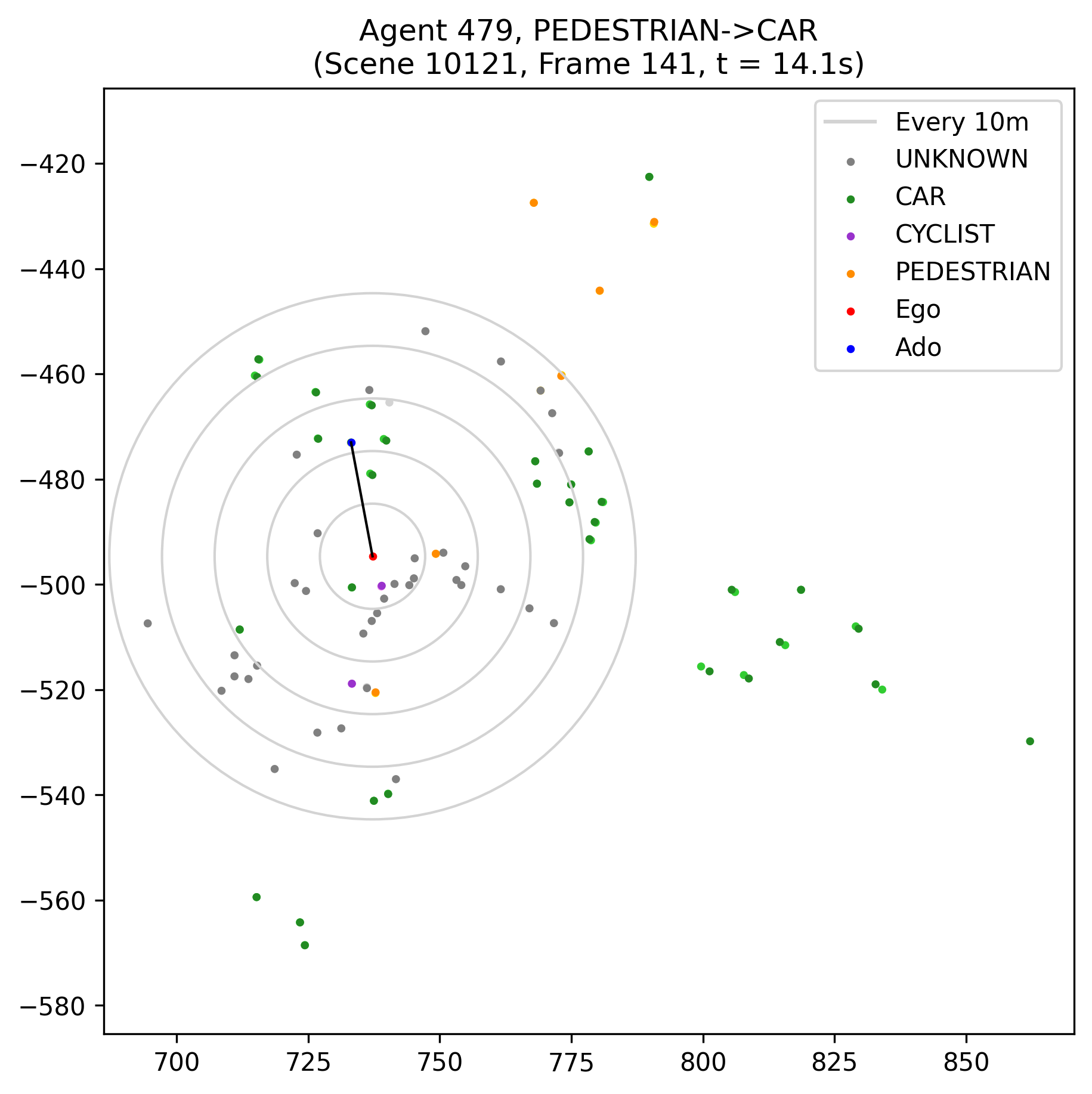}
    \caption{A scenario in the Lyft Level 5 dataset \cite{HoustonZuidhofEtAl2020} where a nearby car is misclassified as a pedestrian in the middle of an intersection. In this example, the misclassified agent (in blue) is only around $20$m away from the ego-vehicle (in red). The solid black line indicates the distance between the two agents and the light gray lines mark $10$m radius increments from the ego-vehicle. The light/dark versions of the other agent colors show the location of the associated agent in the previous frame (light color) and the current frame (dark color). The title indicates the class switch that occurs from the previous to the current frame for the misclassified agent.}
    \label{fig:supp_ped_car_mixup}
\end{figure*}

\begin{figure*}[t]
    \centering
    \includegraphics[width=\linewidth]{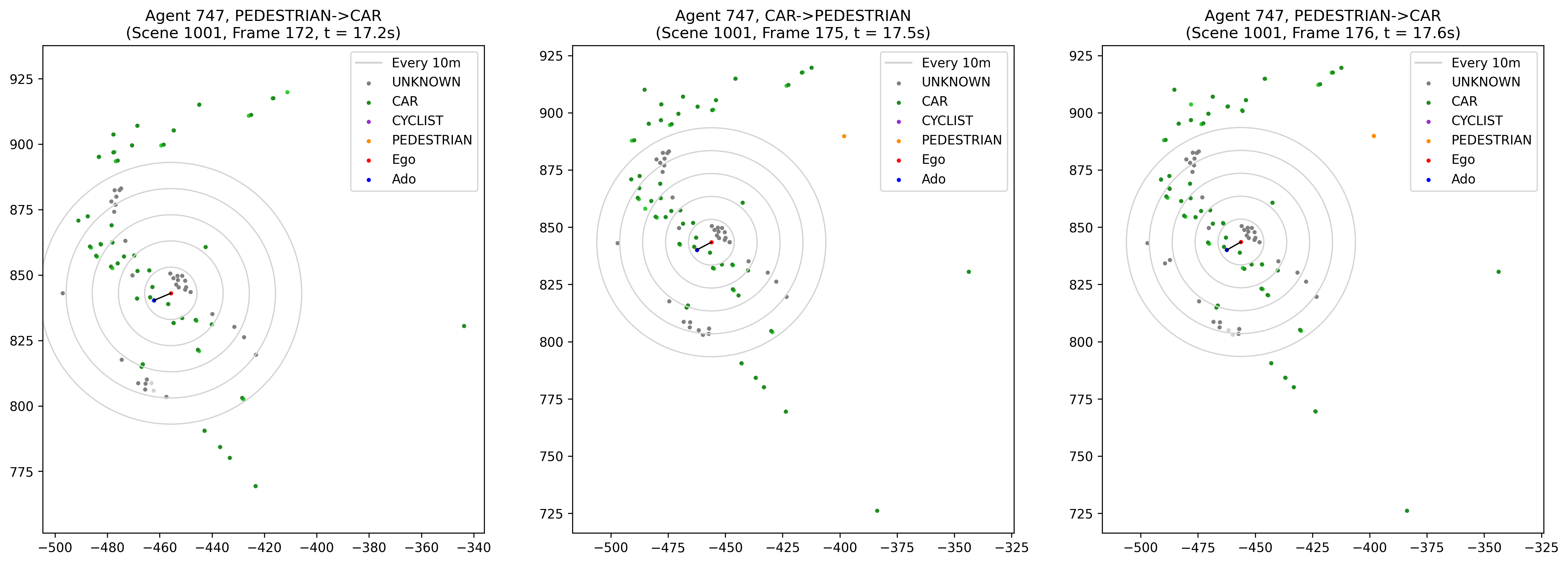}
    \caption{A scenario in the Lyft Level 5 dataset \cite{HoustonZuidhofEtAl2020} where a pedestrian next to the ego-vehicle is misclassified as a car while waiting to cross the street. In this example, the misclassified agent (in blue) is less than $10$m away from the ego-vehicle (in red). The solid black line indicates the distance between the two agents and the light gray lines mark $10$m radius increments from the ego-vehicle. The light/dark versions of the other agent colors show the location of the associated agent in the previous frame (light color) and the current frame (dark color). The title indicates the class switch that occurs from the previous to the current frame for the misclassified agent.}
    \label{fig:supp_ped_near_cars_mixup}
\end{figure*}

\cref{fig:supp_ped_car_mixup} visualizes a scenario where a nearby car is misclassified as a pedestrian in the middle of an intersection. Another example is visualized in \cref{fig:supp_ped_near_cars_mixup} where a pedestrian adjacent to the ego-vehicle is misclassified as a car while waiting to cross the street, demonstrating that class switching is not solely due to an agent being very far away from sensor view.

\subsubsection{Types of Class Switching} 
\begin{figure*}[t]
    \centering
    \includegraphics[width=\linewidth]{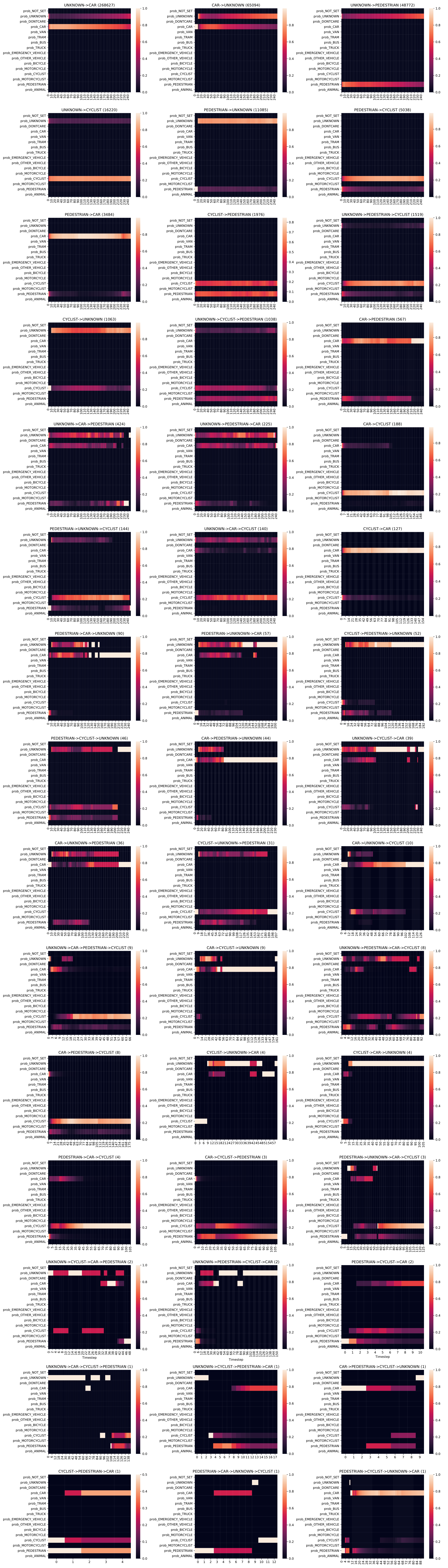}
    \caption{The 15 most common class switches (out of 45 total) in the Lyft Level 5 dataset \cite{HoustonZuidhofEtAl2020}, as well as the mean agent class probabilities over time for each case. Each subfigure title indicates the type of class switch as well as the number of affected agents in brackets.}
    \label{fig:supp_top_15_mixups}
\end{figure*}

\cref{fig:supp_top_15_mixups} visualizes the 15 most common class switches (out of 45 total) in the Lyft dataset, as well as the mean agent class probabilities for each case. As can be seen, there are thousands of agents which experience known-class to known-class switches (e.g., pedestrian to car).

\subsubsection{Probability of the Most-Likely Class}
\label{supp:lyft_prob_ml_class} 
\begin{figure*}[t]
    \centering
    \includegraphics[width=\linewidth]{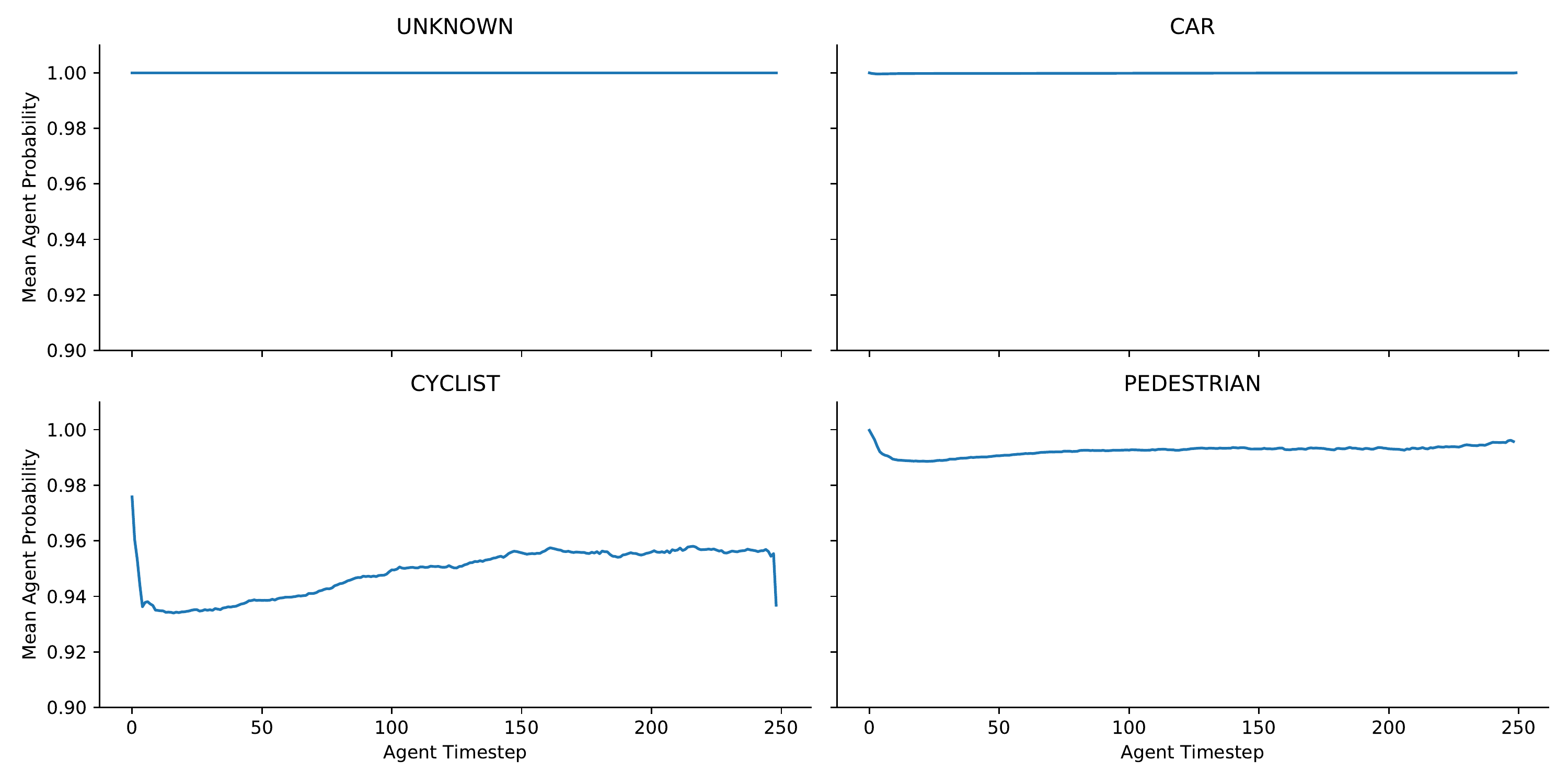}
    \caption{The average probability of the agent's most-likely class over time for each class in the Lyft Level 5 dataset \cite{HoustonZuidhofEtAl2020}.}
    \label{fig:supp_agent_type_prob_vs_age}
\end{figure*}

\cref{fig:supp_agent_type_prob_vs_age} shows the average probability of the agent's most-likely class (indicated by the subfigure title). As can be seen, the most-likely class has more than $92\%$ probability on average per agent timestep.

\subsection{Detailed \dataname{} Dataset Statistics}
\label{supp:detailed_new_statistics}

Note that when collecting statistics or evaluating methods that require agent classes (e.g., Trajectron++ and MATS) on the \dataname{} dataset, we use an agent's most often most-likely class as their fixed classification.


\subsection{Additional Results}
\label{supp:additional_results}

\subsubsection{Lyft Dataset}
\label{supp:additional_lyft_results}
\cref{tab:supp_lyft} shows the per-class performance of our method, its ablations, and baselines on the Lyft dataset. As in \cref{tab:lyft_quant}, our method generally performs the best across all classes and is similar to the one-hot ablation due to the abundance of one-hot class probabilities in the Lyft dataset.

\begin{table*}[t]
\centering
\caption{Per-class performance on the Lyft Level 5 dataset \cite{HoustonZuidhofEtAl2020}. Lower is better, bold is best.}
\label{tab:supp_lyft}
\begin{tabular}{l|c|ccc|c|ccc}
\toprule
\multicolumn{1}{c|}{\textbf{Lyft -- Car}} & \multicolumn{1}{c}{\textbf{ADE}} & \multicolumn{3}{c|}{\textbf{FDE (m)}} & \multicolumn{1}{c}{\textbf{ANLL}} & \multicolumn{3}{c}{\textbf{FNLL (nats)}} \\ \cline{1-9} 
\multicolumn{1}{c|}{\textbf{Pred. Horizon}} & 3s & 1s      & 2s        & 3s & 3s & 1s      & 2s        & 3s\TBstrut \\ \midrule
MATS \cite{IvanovicElhafsiEtAl2020} & $2.33$ & $0.46$ & $2.21$ & $6.17$ & $4.75$ & $0.85$ & $4.10$ & $13.67$\\
Trajectron++ \cite{SalzmannIvanovicEtAl2020}  & $0.77$ & $0.43$ & $0.97$ & $1.74$ & $-0.19$ & $-0.63$ & $0.75$ & $1.71$\\
Multi-Head \cite{djuric2020multinet} & $0.80$ & $0.43$ & $1.00$ & $1.81$ & $-0.20$ & $-0.63$ & $0.73$ & $1.71$ \\
One-Hot & $0.76$ & $0.42$ & $0.95$ & $\mathbf{1.71}$ & $\mathbf{-0.66}$ & $\mathbf{-0.91}$ & $\mathbf{0.24}$ & $\mathbf{1.09}$\\
\hline
\algname{} (ours)        & $\mathbf{0.75}$ & $\mathbf{0.40}$ & $\mathbf{0.94}$ & $\mathbf{1.71}$ & $-0.64$ & $\mathbf{-0.91}$ & $0.27$ & $1.17$ \Tstrut{} \\
\bottomrule

\toprule
\multicolumn{1}{c|}{\textbf{Lyft -- Cyclist}} & \multicolumn{1}{c}{\textbf{ADE}} & \multicolumn{3}{c|}{\textbf{FDE (m)}} & \multicolumn{1}{c}{\textbf{ANLL}} & \multicolumn{3}{c}{\textbf{FNLL (nats)}} \\ \cline{1-9} 
\multicolumn{1}{c|}{\textbf{Pred. Horizon}} & 3s & 1s      & 2s        & 3s & 3s & 1s      & 2s        & 3s\TBstrut \\ \midrule
MATS \cite{IvanovicElhafsiEtAl2020} & $1.14$ & $0.33$ & $1.25$ & $2.45$ & $3.20$ & $0.16$ & $2.59$ & $11.14$\\
Trajectron++ \cite{SalzmannIvanovicEtAl2020}  & $0.64$ & $0.29$ & $0.81$ & $1.58$ & $-0.45$ & $-0.93$ & $0.66$ & $1.70$\\
Multi-Head \cite{djuric2020multinet} & $0.55$ & $0.28$ & $0.71$ & $1.26$ & $-0.67$ & $-1.08$ & $0.38$ & $1.33$ \\
One-Hot & $0.51$ & $0.27$ & $0.66$ & $1.17$ & $\mathbf{-1.07}$ & $-1.34$ & $\mathbf{-0.04}$ & $\mathbf{0.78}$\\
\hline
\algname{}  (ours)   & $\mathbf{0.48}$ & $\mathbf{0.25}$ & $\mathbf{0.62}$ & $\mathbf{1.12}$ & $\mathbf{-1.07}$ & $\mathbf{-1.37}$ & $\mathbf{-0.04}$ & $\mathbf{0.78}$ \Tstrut{} \\
\bottomrule

\toprule
\multicolumn{1}{c|}{\textbf{Lyft -- Pedestrian}} & \multicolumn{1}{c}{\textbf{ADE}} & \multicolumn{3}{c|}{\textbf{FDE (m)}} & \multicolumn{1}{c}{\textbf{ANLL}} & \multicolumn{3}{c}{\textbf{FNLL (nats)}} \\ \cline{1-9} 
\multicolumn{1}{c|}{\textbf{Pred. Horizon}} & 3s & 1s      & 2s        & 3s & 3s & 1s      & 2s        & 3s\TBstrut \\ \midrule
MATS \cite{IvanovicElhafsiEtAl2020} & $0.68$ & $0.28$ & $0.70$ & $1.42$ & $3.29$ & $0.06$ & $1.61$ & $10.69$\\
Trajectron++ \cite{SalzmannIvanovicEtAl2020}  & $\mathbf{0.30}$ & $0.18$ & $0.39$ & $\mathbf{0.62}$ & $-1.14$ & $-1.40$ & $-0.09$ & $0.74$\\
Multi-Head \cite{djuric2020multinet} & $\mathbf{0.30}$ & $0.18$ & $0.39$ & $\mathbf{0.62}$ & $\mathbf{-1.36}$ & $\mathbf{-1.64}$ & $\mathbf{-0.30}$ & $\mathbf{0.53}$ \\
One-Hot & $\mathbf{0.30}$ & $\mathbf{0.17}$ & $\mathbf{0.38}$ & $0.63$ & $-1.33$ & $-1.60$ & $-0.24$ & $0.64$\\
\hline
\algname{}  (ours)    & $\mathbf{0.30}$ & $\mathbf{0.17}$ & $\mathbf{0.38}$ & $0.65$ & $-1.31$ & $\mathbf{-1.64}$ & $\mathbf{-0.30}$ & $\mathbf{0.53}$ \Tstrut{} \\
\bottomrule

\toprule
\multicolumn{1}{c|}{\textbf{Lyft -- Unknown}} & \multicolumn{1}{c}{\textbf{ADE}} & \multicolumn{3}{c|}{\textbf{FDE (m)}} & \multicolumn{1}{c}{\textbf{ANLL}} & \multicolumn{3}{c}{\textbf{FNLL (nats)}} \\ \cline{1-9} 
\multicolumn{1}{c|}{\textbf{Pred. Horizon}} & 3s & 1s      & 2s        & 3s & 3s & 1s      & 2s        & 3s\TBstrut \\ \midrule
MATS \cite{IvanovicElhafsiEtAl2020} & $0.71$ & $0.28$ & $0.71$ & $1.57$ & $3.87$ & $-0.09$ & $1.57$ & $12.76$\\
Trajectron++ \cite{SalzmannIvanovicEtAl2020}  & $0.18$ & $\mathbf{0.15}$ & $0.22$ & $0.28$ & $\mathbf{-2.43}$ & $\mathbf{-2.43}$ & $\mathbf{-1.85}$ & $\mathbf{-1.42}$\\
Multi-Head \cite{djuric2020multinet} & $\mathbf{0.17}$ & $\mathbf{0.15}$ & $\mathbf{0.21}$ & $\mathbf{0.27}$ & $-2.38$ & $-2.40$ & $-1.83$ & $-1.39$\\
One-Hot & $0.18$ & $0.16$ & $0.22$ & $0.28$ & $-2.38$ & $-2.38$ & $-1.83$ & $-1.40$\\
\hline
\algname{}  (ours)  & $0.18$ & $\mathbf{0.15}$ & $0.22$ & $0.28$ & $-2.38$ & $-2.38$ & $-1.81$ & $-1.38$ \Tstrut{} \\
\bottomrule
\end{tabular}

\end{table*}

\subsubsection{\dataname{} Dataset}
\label{supp:additional_new_results}
\cref{tab:supp_new} shows the per-class performance of our method, its ablations, and baselines on our \dataname{} dataset. \algname{}'s strong performance across agent classes is evident, and while other baselines or ablations yield strong performance on specific classes, they are not able to maintain performance across agent classes in general. 

Note that we could not evaluate MATS \cite{IvanovicElhafsiEtAl2020} on unknown agents ($0.1\%$ of the data) due to MATS' training-time computational requirements. Specifically, constructing dense square matrices (and backpropagating through them) to model many batched scenes exhausted our computational resources. To remedy this, we temporally subsampled our \dataname{} dataset for MATS, which removed (short-lived) unknown agents.

\subsection{Additional Training Information}
\label{supp:training_info}

All methods were implemented in PyTorch \cite{PaszkeGrossEtAl2017} on a computer running Ubuntu 18.04 containing an AMD Ryzen 1800X CPU and two NVIDIA GTX 1080 Ti GPUs.


We anneal the $\beta$ hyperparameter in \cref{eqn:loss_fn} following an increasing sigmoid \cite{BowmanVilnisEtAl2015}.
Specifically, $\beta$ takes a low value at early training iterations so that the model is encouraged to encode information in $z$. At later training iterations, a higher $\beta$ value shifts the role of information encoding from $q_\phi(z \mid \mathbf{x}, \mathbf{y})$ to $p_\theta(z \mid \mathbf{x}, \hat{\mathbf{c}})$.

To avoid overfitting to environment-specific characteristics, such as the general directions that agents move, we augment the data from each scene with rotation \cite{SchollerAravantinosEtAl2020}. 
In particular, we rotate all trajectories in a scene around the scene’s origin by $\gamma$, where $\gamma$ varies from $0^\circ$ to $360^\circ$ (exclusive) in $15^\circ$ intervals, as in \cite{SalzmannIvanovicEtAl2020}. We apply this augmentation to autonomous driving datasets because most of them are recorded in cities whose streets are roughly orthogonal and separated by blocks. While this augmentation is equivalent to rotating all trajectories to a canonical agent-centric frame, we chose to rotate all trajectories at train-time and train the model to be rotation-invariant since it avoids the need to perform canonical (and possibly noisy) trajectory rotation online at test time.

Finally, the Gumbel-Softmax reparameterization \cite{JangGuEtAl2017} is \emph{not} used to backpropagate through the Categorical latent variable $z$ because $z$ is not sampled during training. Instead, the first term of \cref{eqn:loss_fn} is directly enumerated and summed since the latent space has only $|Z| = 25$ discrete elements. 

\subsection{minADE and minFDE Results}
\label{supp:minADE_minFDE}

\begin{table*}[t]
\centering
\caption{Evaluating with the minADE and minFDE metrics (over 20 samples) shows that \algname{} still outperforms existing approaches in the face of uncertainty (\dataname{} dataset). Further, as expected, performance is similar on the Lyft data (which has virtually no class uncertainty). Lower is better, bold is best. SE = Standard Error.}
\label{tab:supp_minADE_minFDE}
\begin{tabular}{l|c|ccc}
\toprule
\multicolumn{1}{c|}{\textbf{Lyft Level 5} \cite{HoustonZuidhofEtAl2020}} & \multicolumn{1}{c}{\textbf{minADE {\small{$\pm$ SE}}}} & \multicolumn{3}{c}{\textbf{minFDE {\small{$\pm$ SE}} (m)}} \\ \cline{1-5} 
\multicolumn{1}{c|}{\textbf{Pred. Horizon}}                         & 3s & 1s      & 2s        & 3s \TBstrut \\ \midrule
Trajectron++ \cite{SalzmannIvanovicEtAl2020} & $\mathbf{0.23}$\small{$\pm 3$e-$3$} & $0.10$\small{$\pm 1$e-$3$} & $0.23$\small{$\pm 3$e-$3$} & $0.40$\small{$\pm 6$e-$3$}\\
Multi-Head \cite{djuric2020multinet} & $0.24$\small{$\pm 3$e-$3$} & $0.10$\small{$\pm 1$e-$3$} & $0.22$\small{$\pm 3$e-$3$} & $0.38$\small{$\pm 6$e-$3$} \\
One-Hot & $0.26$\small{$\pm 3$e-$3$} & $\mathbf{0.09}$\small{$\pm 1$e-$3$} & $\mathbf{0.20}$\small{$\pm 3$e-$3$} & $\mathbf{0.36}$\small{$\pm 6$e-$3$}\\
\hline
\algname{} (ours) & $0.26$\small{$\pm 3$e-$3$} & $\mathbf{0.09}$\small{$\pm 1$e-$3$} & $0.21$\small{$\pm 3$e-$3$} & $0.38$\small{$\pm 6$e-$3$} \Tstrut{}\\
\bottomrule

\toprule
\multicolumn{1}{c|}{\textbf{PUP}} & \multicolumn{1}{c}{\textbf{minADE {\small{$\pm$ SE}}}} & \multicolumn{3}{c}{\textbf{minFDE {\small{$\pm$ SE}} (m)}} \\ \cline{1-5} 
\multicolumn{1}{c|}{\textbf{Pred. Horizon}}                         & 3s & 1s      & 2s        & 3s \TBstrut \\ \midrule
Trajectron++ \cite{SalzmannIvanovicEtAl2020} & $0.36$\small{$\pm 0.02$} & $0.21$\small{$\pm 0.02$} & $0.37$\small{$\pm 0.02$} & $0.52$\small{$\pm 0.04$}\\
Multi-Head \cite{djuric2020multinet} & $0.37$\small{$\pm 0.01$} & $0.21$\small{$\pm 0.01$} & $0.33$\small{$\pm 0.02$} & $0.53$\small{$\pm 0.03$} \\
One-Hot & $0.40$\small{$\pm 0.02$} & $0.17$\small{$\pm 9$e-$3$} & $\mathbf{0.29}$\small{$\pm 0.02$} & $0.53$\small{$\pm 0.05$}\\
\hline
\algname{} (ours) & $\mathbf{0.35}$\small{$\pm 0.01$} & $\mathbf{0.15}$\small{$\pm 9$e-$3$} & $\mathbf{0.29}$\small{$\pm 0.02$} & $\mathbf{0.48}$\small{$\pm 0.03$} \Tstrut{}\\
\bottomrule
\end{tabular}

\end{table*}

\cref{tab:supp_minADE_minFDE} summarizes evaluation on the Lyft and \dataname{} datasets with the minADE and minFDE metrics (over 20 samples).

\begin{table*}[t]
\centering
\caption{Per-class performance on our \dataname{} dataset. Lower is better, bold is best.}
\label{tab:supp_new}
\begin{tabular}{l|c|ccc|c|ccc}
\toprule
\multicolumn{1}{c|}{\textbf{\dataname{} -- Bicycle}} & \multicolumn{1}{c}{\textbf{ADE}} & \multicolumn{3}{c|}{\textbf{FDE (m)}} & \multicolumn{1}{c}{\textbf{ANLL}} & \multicolumn{3}{c}{\textbf{FNLL (nats)}} \\ \cline{1-9} 
\multicolumn{1}{c|}{\textbf{Pred. Horizon}} & 3s & 1s      & 2s        & 3s & 3s & 1s      & 2s        & 3s\TBstrut \\ \midrule
MATS \cite{IvanovicElhafsiEtAl2020} & $2.33$ & $0.76$ & $2.03$ & $4.20$ & $7.02$ & $3.54$ & $5.81$ & $11.70$\\
Trajectron++ \cite{SalzmannIvanovicEtAl2020}  & $0.61$ & $0.35$ & $0.74$ & $1.37$ & $-0.63$ & $-1.01$ & $-0.08$ & $0.60$\\

Multi-Head \cite{djuric2020multinet} & $0.71$ & $0.45$ & $0.88$ & $1.42$ & $-0.53$ & $-0.95$ & $0.06$ & $0.81$  \\
One-Hot & $0.56$ & $\mathbf{0.32}$ & $0.69$ & $1.23$ & $\mathbf{-0.99}$ & $-1.40$ & $\mathbf{-0.41}$ & $\mathbf{0.31}$\\
\hline
\algname{} (ours) & $\mathbf{0.54}$ & $0.33$ & $\mathbf{0.66}$ & $\mathbf{1.13}$ & $-0.98$ & $\mathbf{-1.41}$ & $-0.38$ & $0.32$ \Tstrut{}\\
\bottomrule

\toprule
\multicolumn{1}{c|}{\textbf{\dataname{} -- Car}} & \multicolumn{1}{c}{\textbf{ADE}} & \multicolumn{3}{c|}{\textbf{FDE (m)}} & \multicolumn{1}{c}{\textbf{ANLL}} & \multicolumn{3}{c}{\textbf{FNLL (nats)}} \\ \cline{1-9} 
\multicolumn{1}{c|}{\textbf{Pred. Horizon}} & 3s & 1s      & 2s        & 3s & 3s & 1s      & 2s        & 3s\TBstrut \\ \midrule
MATS \cite{IvanovicElhafsiEtAl2020} & $1.01$ & $0.42$ & $0.89$ & $1.73$ & $5.89$ & $1.57$ & $2.63$ & $13.47$ \\
Trajectron++ \cite{SalzmannIvanovicEtAl2020}  & $0.57$ & $0.37$ & $0.68$ & $1.12$ & $-1.10$ & $-1.39$ & $-0.73$ & $-0.14$ \\
Multi-Head \cite{djuric2020multinet} & $0.70$ & $0.50$ & $0.86$ & $1.31$ & $-0.98$ & $-1.34$ & $-0.55$ & $0.15$  \\
One-Hot & $0.55$ & $\mathbf{0.36}$ & $\mathbf{0.65}$ & $1.10$ & $-1.29$ & $-1.58$ & $-0.90$ & $-0.27$\\
\hline
\algname{} (ours) & $\mathbf{0.54}$ & $\mathbf{0.36}$ & $\mathbf{0.65}$ & $\mathbf{1.09}$ & $\mathbf{-1.31}$ & $\mathbf{-1.60}$ & $\mathbf{-0.92}$ & $\mathbf{-0.29}$ \Tstrut{}\\
\bottomrule

\toprule
\multicolumn{1}{c|}{\textbf{\dataname{} -- Largevehicle}} & \multicolumn{1}{c}{\textbf{ADE}} & \multicolumn{3}{c|}{\textbf{FDE (m)}} & \multicolumn{1}{c}{\textbf{ANLL}} & \multicolumn{3}{c}{\textbf{FNLL (nats)}} \\ \cline{1-9} 
\multicolumn{1}{c|}{\textbf{Pred. Horizon}} & 3s & 1s      & 2s        & 3s & 3s & 1s      & 2s        & 3s\TBstrut \\ \midrule
MATS \cite{IvanovicElhafsiEtAl2020} & $1.42$ & $\mathbf{0.62}$ & $1.27$ & $2.39$ & $8.77$ & $5.02$ & $5.83$ & $15.46$\\
Trajectron++ \cite{SalzmannIvanovicEtAl2020}  & $0.88$ & $0.66$ & $1.04$ & $1.54$ & $0.31$ & $-0.03$ & $0.65$ & $1.27$ \\
Multi-Head \cite{djuric2020multinet} & $1.07$ & $0.83$ & $1.28$ & $1.80$ & $0.42$ & $0.03$ & $0.82$ & $1.51$  \\
One-Hot & $\mathbf{0.85}$ & $\mathbf{0.62}$ & $\mathbf{1.00}$ & $\mathbf{1.53}$ & $0.03$ & $-0.27$ & $0.43$ & $\mathbf{1.07}$\\
\hline
\algname{} (ours) & $0.88$ & $0.63$ & $1.03$ & $1.60$ & $\mathbf{0.02}$ & $\mathbf{-0.29}$ & $\mathbf{0.42}$ & $\mathbf{1.07}$ \Tstrut{}\\
\bottomrule

\toprule
\multicolumn{1}{c|}{\textbf{\dataname{} -- Motorcycle}} & \multicolumn{1}{c}{\textbf{ADE}} & \multicolumn{3}{c|}{\textbf{FDE (m)}} & \multicolumn{1}{c}{\textbf{ANLL}} & \multicolumn{3}{c}{\textbf{FNLL (nats)}} \\ \cline{1-9} 
\multicolumn{1}{c|}{\textbf{Pred. Horizon}} & 3s & 1s      & 2s        & 3s & 3s & 1s      & 2s        & 3s\TBstrut \\ \midrule
MATS \cite{IvanovicElhafsiEtAl2020} & $0.59$ & $0.28$ & $\mathbf{0.51}$ & $0.98$ & $4.14$ & $0.74$ & $1.34$ & $10.34$\\
Trajectron++ \cite{SalzmannIvanovicEtAl2020}  & $0.52$ & $0.36$ & $0.64$ & $0.96$ & $-0.57$ & $-0.98$ & $-0.10$ & $0.49$\\
Multi-Head \cite{djuric2020multinet} & $0.53$ & $0.34$ & $0.63$ & $1.05$ & $-0.43$ & $-0.92$ & $0.13$ & $0.92$  \\
One-Hot & $\mathbf{0.43}$ & $\mathbf{0.27}$ & $\mathbf{0.51}$ & $\mathbf{0.86}$ & $\mathbf{-1.02}$ & $\mathbf{-1.41}$ & $\mathbf{-0.52}$ & $\mathbf{0.07}$\\
\hline
\algname{} (ours) & $0.47$ & $0.29$ & $0.56$ & $0.98$ & $-0.94$ & $-1.36$ & $-0.43$ & $0.21$ \Tstrut{}\\
\bottomrule

\toprule
\multicolumn{1}{c|}{\textbf{\dataname{} -- Pedestrian}} & \multicolumn{1}{c}{\textbf{ADE}} & \multicolumn{3}{c|}{\textbf{FDE (m)}} & \multicolumn{1}{c}{\textbf{ANLL}} & \multicolumn{3}{c}{\textbf{FNLL (nats)}} \\ \cline{1-9} 
\multicolumn{1}{c|}{\textbf{Pred. Horizon}} & 3s & 1s      & 2s        & 3s & 3s & 1s      & 2s        & 3s\TBstrut \\ \midrule
MATS \cite{IvanovicElhafsiEtAl2020} & $0.79$ & $\mathbf{0.32}$ & $0.72$ & $1.32$ & $2.47$ & $0.27$ & $2.35$ & $4.79$\\
Trajectron++ \cite{SalzmannIvanovicEtAl2020}  & $\mathbf{0.50}$ & $\mathbf{0.32}$ & $\mathbf{0.62}$ & $\mathbf{0.97}$ & $\mathbf{0.36}$ & $\mathbf{-0.17}$ & $\mathbf{1.13}$ & $\mathbf{2.05}$\\
Multi-Head \cite{djuric2020multinet} & $0.65$ & $0.41$ & $0.82$ & $1.29$ & $0.66$ & $0.14$ & $1.45$ & $2.31$  \\
One-Hot & $0.58$ & $0.36$ & $0.72$ & $1.18$ & $0.41$ & $-0.11$ & $1.18$ & $2.06$\\
\hline
\algname{} (ours) & $0.62$ & $0.38$ & $0.77$ & $1.24$ & $0.48$ & $-0.04$ & $1.25$ & $2.15$ \Tstrut{}\\
\bottomrule

\toprule
\multicolumn{1}{c|}{\textbf{\dataname{} -- Unknown}} & \multicolumn{1}{c}{\textbf{ADE}} & \multicolumn{3}{c|}{\textbf{FDE (m)}} & \multicolumn{1}{c}{\textbf{ANLL}} & \multicolumn{3}{c}{\textbf{FNLL (nats)}} \\ \cline{1-9} 
\multicolumn{1}{c|}{\textbf{Pred. Horizon}} & 3s & 1s      & 2s        & 3s & 3s & 1s      & 2s        & 3s\TBstrut \\ \midrule
Trajectron++ \cite{SalzmannIvanovicEtAl2020}  & $1.44$ & $0.79$ & $1.86$ & $3.14$ & $1.89$ & $1.80$ & $3.28$ & $3.65$\\
Multi-Head \cite{djuric2020multinet} & $1.41$ & $0.79$ & $1.83$ & $2.97$ & $1.96$ & $1.32$ & $3.28$ & $3.74$  \\
One-Hot & $1.16$ & $0.59$ & $1.51$ & $2.58$ & $1.46$ & $1.25$ & $2.86$ & $3.26$\\
\hline
\algname{} (ours) & $\mathbf{0.86}$ & $\mathbf{0.40}$ & $\mathbf{1.09}$ & $\mathbf{2.10}$ & $\mathbf{0.64}$ & $\mathbf{0.07}$ & $\mathbf{2.00}$ & $\mathbf{2.32}$ \Tstrut{}\\
\bottomrule
\end{tabular}

\end{table*}

\clearpage

\end{document}